# Shapley Computations Using Surrogate Model-Based Trees

Zhipu Zhou[1], Jie Chen, Linwei Hu[2]

June 20, 2022

**Abstract**

Shapley-related techniques have gained attention as both global and local interpretation tools because of their desirable properties. However, their computation using conditional expectations is computationally expensive. Approximation methods suggested in the literature have limitations. This paper proposes the use of a surrogate model-based tree to compute Shapley and SHAP values based on conditional expectation. Simulation studies show that the proposed algorithm provides improvements in accuracy, unifies global Shapley and SHAP interpretation, and the thresholding method provides a way to trade-off running time and accuracy.

## 1 Introduction

The ability to correctly interpret a model and its predictions are important in many regulated sectors such as banking. One aspect of model interpretation is to understand the relative importance of its input features. Shapley value, a concept from cooperative game theory, has been found useful in this context, as it provides unique feature importance attribution and satisfies desirable axioms (Shapley, 1952). Feature attributions can be done using global Shapley, which measures the overall importance of a predictor (Owen, 2014; Song, et al., 2016) or SHAP (SHapley Additive exPlanation) by (Lundberg & Lee, 2017) that explains individual prediction by computing the contribution of each feature to that prediction.

For global Shapley, (Owen, 2014; Song, et al., 2016) propose the use of the conditional expectation in calculating the contributions. The approach measures the explanatory power of a subset of input features while other features are marginalized out by their conditional distribution. Conditional expectation is also used in Shapley-Owen Index to interpret feature interactions (Rabitti & Borgonovo, 2019). For SHAP, (Lundberg & Lee, 2017) also proposes the use of conditional expectations. However, the exact computation of conditional expectation is difficult, and they propose the use of Kernel SHAP as an approximation which assumes the features are independent thus this is equivalent to using marginal distributions.

---

[1] Email: Lucas.Zhou@wellsfargo.com
[2] The views expressed in the paper are those of the authors and do not represent the views of Wells Fargo.

There has been extensive discussion in the literature as to whether one should use conditional or marginal distributions to calculate Shapley-related values, see (Chen, et al., 2020; Janzing, et al., 2019; Frye, et al., 2021; Sundararajan & Najmi, 2020). (Chen, et al., 2020) suggests that the choice depends on the application. The use of marginal distributions captures the behavior of a particular model. On the other hand, the use of conditional distributions considers the dependencies (correlations) in the input data. (Frye, et al., 2021) argues that using conditional distributions can give a variable nonzero importance measure even if it was not in the data set at the time the model was trained (Mase, et al., 2020), while the use of marginal distributions can hide implicit model dependence on sensitive features.

As an alternative to the approximation using marginal distributions, (Lundberg, et al., 2019) proposes Tree SHAP, which uses a tree structure to approximate the conditional expectation. In each splitting node of the tree, the splitting variable is assumed to be independent of other feature variables (a problem called *path dependency issue* in the present paper). (Aas, et al., 2020) proposes other methods to compute conditional expectation, including multivariate Gaussian distribution assumption, Gaussian copula, and empirical conditional distribution. Non-parametric methods such as empirical conditional distribution and nearest neighbor method (Broto, et al., 2020), and other methods such as Cohort Shapley (Mase, et al., 2020), require computing the distance among all observations and for all subsets of features, and thus are still computationally expensive.

In this paper, we propose the use of a surrogate model-based tree to compute the conditional expectation for both global Shapley and SHAP. We make the following contributions:
1. From the theoretical perspective, we propose to use SLIM (Surrogate Locally Interpretable Model) to compute the conditional expectation from the observational data. SLIM is a model-based tree algorithm that conducts supervised partitioning of the space of predictors (Hu, et al., 2022 (to appear)). We discover that in the existing tree related interpretation model such as Tree SHAP (Lundberg, et al., 2019), a splitting variable (or a path) is assumed to be independent of other variables. Such an assumption does not hold in general. To our knowledge, there is no investigation in the literature about how to solve the path dependency issue. We offer a machine learning model-based method to resolve the path dependency issue. We demonstrate empirically on several case studies that, compared with several major methods exist in the literature, our proposed method provides remarkable improvement in the computational accuracy.

2. From the methodological perspective, our proposed method can provide both the global Shapley and the SHAP values. For the current SHAP methods, the local feature importance measures are aggregated to a global importance score in an ad-hoc approach. Such an approach is sensitive to the aggregation method been used. Our work unifies both global and local measures in a consistent theoretical and implementation framework. Such a consistency is helpful to understand and explain two types of Shapley results.

3. From the computationally efficient perspective, we provide the thresholding method that selects parts of all subsets of features by the size of each subset. Original Shapley formulation requires to go through all the subsets of features. It will be computational expensive when feature size is large. Thresholding method offers the user an option to pursuit computational efficiency at the cost of reducing the result accuracy.

The rest of this paper is organized as follows. Section 2 introduces the related work, including Tree SHAP and its path dependency issue, and a brief introduction to SLIM. Section 3 discusses the proposed algorithm in detail. Section 4 illustrates the implementation of the proposed method and the comparison of numerical results for case studies. Section 5 summarizes our work.

## 2 Related work

### 2.1 Definition of Shapley and SHAP values

Consider a prediction model $\hat{f}(X)$ with a set of $p$ input features $\mathcal{K} = \{1, 2, \dots, p\}$. Given a defined value function $c: 2^{\mathcal{K}} \mapsto \mathbb{R}$ with $c(\emptyset) = 0$, the Shapley value of feature $X_i$ with respect to $c(.)$ is defined as

$$\phi_i = \sum_{u \subseteq \mathcal{K} \setminus \{i\}} \frac{(p - |u| - 1)! \, |u|!}{p!} [c(\{i\} \cup u) - c(u)]$$

where $|u|$ indicates the size of subset $u \subseteq \mathcal{K}$. For global Shapley, (Owen, 2014; Song, et al., 2016) propose a convenient way to measure the global interpretation via:

$$c(u) = Var(\mathbb{E}(\hat{f}(X)|X_u)) \quad (1)$$

where features not in $u$ are marginalizing out via their conditional distribution.

For SHAP, (Lundberg & Lee, 2017) proposes to measure local interpretation via

$$c(u) = \mathbb{E}(\hat{f}(X)|X_u). \quad (2)$$

Both choices of the value function requires computing conditional expectations. In practice, for a given subset $u$, it is generally challenging to compute $\mathbb{E}(\hat{f}(X)|X_u)$ from the observational data $X$ and the fitted response $\hat{Y} = \hat{f}(X)$ as the conditional distribution is difficult to know and the true relationship between $X_u$ and $\hat{Y}$ may be complex.

## 2.2 Tree SHAP and path dependency issue

Tree SHAP is an algorithm for approximating the exact condition-expectation based local feature importance for tree ensemble methods such as gradient boosting trees and random forests (Lundberg, et al., 2019). Given a prediction at a particular value in the predictor space ("instance"), Tree SHAP computes the product of the constant response fit in a terminal node and the probability that the instance falls into that terminal node and computes the sum of the products from all terminal nodes as the value of the conditional expectation. We call the probability that the instance is falling into a terminal node as the path probability.

In Tree SHAP, path probability is computed by implicitly assuming that splitting variables are independent of other remaining variables. Such an assumption does not hold in general, and we refer to this issue as *path dependency problem.* In the later part of this paper, we discuss a solution to address this problem.

## 2.3 Surrogate Tree-Based Model (SLIM)

SLIM refers to a surrogate model-based tree developed in (Hu, et al., 2022 (to appear)) for interpreting complex machine learning model algorithms. The idea is to treat the output of a machine learning (ML) algorithm as response and fit a single model-based tree to this pseudo-response. They recommend that use of GAM (Generalized Additive Models) at each node to capture the nonlinearities and use the splits in the trees to capture the interactions. They show that a single tree is stable when fit to the outputs of the ML algorithm (as it smooths the noise) and the fitted tree leads to good interpretations. Readers are referred to (Hu, et al., 2022 (to appear)) for more details.

## 3 Algorithm

### 3.1 General flow of the algorithm

We will introduce the main steps of the algorithm first and discuss the details of the algorithm later. The algorithm consists of six main steps:

1. Build and tune a SLIM tree. Section 2.3 provides a brief introduction to a SLIM tree.

2. In each splitting node of the tree, choose a smaller subset of top important variables for the splitting variable. Section 3.2.2 provides more details.
3. In each splitting node of the tree, build machine learning models for all possible combinations of top important variables for the purpose of path probability calculation. Save all models for later use. Section 3.2.2 provides more details.
4. Select significant subsets of features via thresholding method. Please refer to Section 3.3.2.
5. For each selected feature subset, compute its value function. Specifically, for each terminal node, compute the path probability (using models fitted from Step 3) and the local conditional expectation, and compute their product. The sum of the products of all the terminal nodes will be taken as the value function of a particular feature subset. This step involves materials in Section 3.2.2 and 3.2.3.
6. Compute the global Shapley or SHAP values by aggregating all the value functions. Please refer to Section 3.3.1 for the aggregation by the weighted least squares (WLS) formulation.

In the following sections, we will start with a motivating example and discuss the algorithm in detail.

## 3.2 Computing conditional expectations

### 3.2.1 A motivating example

Suppose there are 5 input features $X_1, X_2, X_3, X_4, X_5$, and a model-based tree is fit to the outputs of a ML model $\hat{Y} = \hat{f}(X)$ as shown in Figure 1.

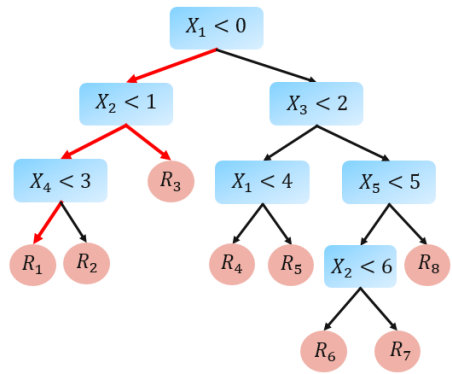

*Figure 1: A fitted model-based tree between X and the fitted response $\hat{Y}$. A blue rectangle represents a splitting node, an orange circle represents a terminal node. The paths highlighted by red arrows are the possible paths for the instance $x = (-2,1,3,0,1)$ with the feature subset $u = \{1, 4\}$.*

Here $R_m$ represents the event that an observational instance $x$ is partitioned into the $m$-th terminal node. Further suppose that a given subset is $u = \{1,4\}$ and one observational instance is $x = (-2,1,3,0,1)$. Since $1 \in u$ and $x_1 = -2$, the root splitting node $\{X_1 < 0\}$ allocates $x$ to its left child node. At the splitting node $\{X_2 < 1\}$, because $\{2\} \notin u$, the instance $x$ has probabilities to go to either the left child node or the right child node. At the next splitting node $\{X_4 < 3\}$, because $\{4\} \in u$ and $x_4 = 0$, $x$ goes to the terminal node $R_1$. Therefore, $x$ can fall into two terminal nodes: $R_1$ and $R_3$. Based on the *law of total expectation* (Weiss, 2005):

$$\mathbb{E}(\hat{Y}|X_u = x_u) = \mathbb{E}\left(\hat{Y}\middle|X_u = (-2,0)\right) = \sum_{i=1,3} p(R_i|X_u = (-2,0))\mathbb{E}(\hat{Y}|R_i, X_u = (-2,0))$$

This equation implies that it is only necessary to compute two quantities: 1) the conditional probability of an instance falling to each terminal node, and 2) the fitted value of $\hat{Y}$ that is locally in each terminal node. While directly computing $\mathbb{E}(\hat{Y}|X_u = x_u)$ is challenging since the global relationship between $X_u$ and $\hat{Y}$ can be complex, locally a simpler, more interpretable model can be fitted in each region $R_i$ to compute the local conditional expectation $\mathbb{E}(\hat{Y}|R_i, X_u = (-2,0))$. In a SLIM tree, the local conditional expectation is computed by a GAM that can account for potential non-linearity.

To generalize, given a prediction model $\hat{f}(X)$ with $p$ features $X$ and the predictions $\hat{Y}$, SLIM can be used to compute the conditional expectation $\mathbb{E}(\hat{Y}|X_u)$ as follows:

$$\mathbb{E}(\hat{Y}|X_u) = \sum_{m=1}^{M} p(R_m|X_u)\mathbb{E}(\hat{Y}|R_m, X_u)$$

where $M$ is the number of terminal nodes in SLIM. Two quantities needed for the computation are the path probability $p(R_m|X_u)$ and the local conditional expectation $\mathbb{E}(\hat{Y}|R_m, X_u)$. We will separately discuss their computation in detail.

### 3.2.2 Computing the path probabilities

Suppose that in the SLIM tree, there are $K$ splitting conditions (splitting events) along the path to the $m$-th terminal node $R_m$. The $K$ splitting conditions are denoted as $A_1, A_2, \ldots, A_K$. For example, $A_1$ may express the event $\{X_5 > 0\}$. As Figure 2 illustrates, an observational instance $x$ will fall into the $m$-th terminal node $R_m$ if and only if all the splitting events along the path of $R_m$ occur. Therefore, by the chain rule of conditional probability, we have

$$\begin{aligned}p(R_m|X_u = x_u) &= p(\cap_{i=1}^{K} A_i | X_u = x_u) \\ &= p(A_1|X_u = x_u)p(A_2|A_1, X_u = x_u) \ldots p(A_i|\cap_{j<i} A_j, X_u = x_u) \\ &\ldots p(A_K|\cap_{j<K} A_j, X_u = x_u)\end{aligned}$$

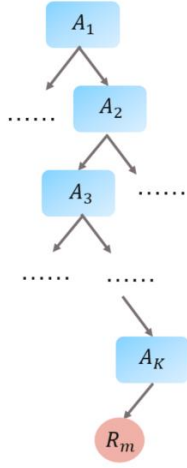

Figure 2: An instance $x \in R_m$ if and only if all the corresponding splitting events along the path to $R_m$ occur.

The computation of the path probability turns out to be the computation of individual conditional probability that the $i$-th splitting event occurs, conditional on the instance $x$ and all the previous splitting events have occurred. There are two cases in the computation of each individual conditional probability $p(A_i|\cap_{j<i} A_j, X_u = x_u)$:

1. **Case 1**: the splitting variable of the event $A_i$ is in the subset $u$. In this case, if the instance $x$ falls into the region $R_m$, then the event $A_i$ will occur with probability 1, conditional on all previous splitting events occur. That is,

$$p(A_i|X_u = x_u) = 1 \text{ for } i = 1 \text{ and all } x \in R_m$$

or

$$p(A_i|\cap_{j<i} A_j, X_u = x_u) = 1 \text{ for } i > 1 \text{ and all } x \in R_m$$

2. **Case 2:** the splitting variable of the event $A_i$ is not in the subset $u$. In this case, features in the subset $u$ may have an impact on whether the event $A_i$ will occur due to the dependency between the splitting variable and the variables in $u$. However, in the current literature such as Tree SHAP, the splitting variable is assumed to be independent of all the other variables, and the conditional probability is computed as

$$p(A_i|\cap_{j<i} A_j, X_u = x_u) = p(A_i|\cap_{j<i} A_j) = \frac{|\{x|x \in A_{i+1}\}|}{|\{x|x \in A_i\}|} \quad (3)$$

where $|\{x|x \in A_i\}|$ denotes the number of instances $x$ that falls into the splitting node $A_i$ when the tree is fitted. Generally, in many realistic applications, such an independent assumption is not held especially for highly correlated features, which we describe as the path dependency issue.

To our knowledge, there is no discussion in the literature discussing or investigating the path dependency issue. We propose to build a ML model to locally estimate $p(A_i|\cap_{j<i} A_j, X_u = x_u)$. The ML model can be a single classification tree or an ensemble of trees (e.g., random forest). In the machine learning model, an indicator function of the splitting event $I_{(A_i)}$ is taken as the binary response variable, and all features that are in the feature subset $u$ are taken as the predictor variables (i.e., $X_i$ for all $i \in u$). Thus, the individual conditional probability is computed as:

$$p(A_i|\cap_{j<i} A_j, X_u) = \mathbb{E}(I_{(A_i)}|\cap_{j<i} A_j, X_u)$$

and the data used to fit the machine learning model would be all the observational instances in the set[3]

$$\{x|x \in \cap_{j<i} A_j\}$$

After computing each of the individual conditional probability, the path probability $p(R_m|X_u = x_u)$ can be computed by the chain rule.

In practice, some subsets $u$ may contain a large number of features while only a small number of them are strongly correlated with the splitting variable, thus building a machine learning model in each splitting node and for each subset $u$ may be unnecessarily time-consuming. A computationally efficient option is designed by choosing a smaller subset $v \subseteq \mathcal{K}$ for each splitting node such that the size of the subset $|v|$ is small (e.g., $|v| = 3$), and the features in $v$ are important variables (e.g., strongly correlated) to the splitting event $I_{(A_i)}$. In each splitting node, $2^{|v|} - 1$ number of machine learning models are built where each model takes $I_{(A_i)}$ as the binary response variable and features in $v'$ are the predictor variables for $\emptyset \neq v' \subseteq v$. Therefore, if $T$ denotes the total number of splitting nodes of a tree, and the size of $v$ is chosen as the same for all splitting nodes, then the total number of machine learning models need to be built to resolve the path dependency issue is $T(2^{|v|} - 1 + 1)$ where the latter 1 represents the model for selecting the important variables of a splitting event. Because SLIM leads to a shallow tree with only a few nodes, $T$ is small. As $|v|$ is chosen to be small, the total number of machine learning models is generally not large. For any subset $u$, there are two cases:

---

[3] There exists an implicit assumption. The ideal case should be the all the data points in the set $\{x|x_u \in \cap_{j<i} A_j\}$. Here we ignore the subset $u$ in the condition.

1. If $u \cap v \neq \emptyset$, then the machine learning model having $u \cap v$ as the predictors will be used to compute the individual conditional probability $p(A_i | \cap_{j<i} A_j, X_u)$.
2. If $u \cap v = \emptyset$, then the individual conditional probability is computed empirically via Equation (3).

### 3.2.3 Computation of local conditional expectation

SLIM interprets the complex prediction model $\hat{f}(X)$ by a set of simpler GAMs such as simple linear models, linear B-spline models, etc. In each region of the SLIM, we make the assumption that the local conditional expectation has the generalized additive form

$$\mathbb{E}(\hat{Y}|R_m, X_u) = g_0 + g_1(X_{k_1}) + g_2(X_{k_2}) + \cdots + g_{|u|}(X_{k_{|u|}})$$

where $k_1, k_2, \ldots, k_{|u|} \in u$ and $g_0 \in \mathbb{R}$ is a constant term. In the case of a simple linear model, $g_i(X_{k_i}) = \beta_i X_{k_i}$. In the case of a linear B-spline model (Hastie, et al., 2008), $g_i(X_{k_i})$ is assumed to be non-linear. In the implementation, linear B-spline models are used as they can account for potential non-linearity. Suppose that in the $m$-th region $R_m$, the number of samples having positive path probability $p(R_m|X_u = x_u^{(i)})$ to this region is $n = n(m)$. Let $x^{(i)}$ and $\hat{y}^{(i)}$ represent the $i$-th sample and the fitted response in this region. For a given feature subset $u \subseteq \mathcal{K}$, these samples form a $n \times |u|$ data matrix $\mathcal{X}_u = \left(x_u^{(1)}, x_u^{(2)}, \ldots, x_u^{(n)}\right)^T$ and the corresponding fitted response vector $\mathcal{Y} = (\hat{y}^{(1)}, \hat{y}^{(2)}, \ldots, \hat{y}^{(n)})^T$. In the linear B-spline model, columns of $\mathcal{X}_u$ are expanded via the linear B-spline transformation using a given number of knots. For the notational convenience, we use $\mathcal{X}_u$ to denote the original $n \times |u|$ data matrix in the simple linear model and denote the data matrix after the linear B-spline transformation in the linear B-spline model. We also denote the $i$-th element of $\mathcal{Y}$ as $\mathcal{Y}^{(i)}$ and the $i$-th row of $\mathcal{X}_u$ as $\mathcal{X}_u^{(i)}$.

As the $i$-th sample $x^{(i)}$ falls into the $m$-th region with a path probability $p(R_m|X_u = x_u^{(i)})$, when estimating the GAM parameters $\beta$ using the least squared method, the quadratic term $\left(\mathcal{Y}^{(i)} - \beta \mathcal{X}_u^{(i)}\right)^2$ would be computed with the path probability of the $i$-th sample. Therefore, if $W = diag(\{w_i\})$ denotes a $n \times n$ diagonal matrix with the $i$-th diagonal element to be $w_i = p(R_m|X_u = x_u^{(i)})$, then the parameters of the GAM should be estimated via the weighted least squares

$$\min_{\beta} \sum_{i=1}^{n} w_i \left(\mathcal{Y}^{(i)} - \beta \mathcal{X}_u^{(i)}\right)^2$$

and the regression coefficients are computed as

$$\hat{\beta} = (\mathcal{X}_u^T W \mathcal{X}_u)^{-1} \mathcal{X}_u^T W \mathcal{Y}$$

and the local conditional expectation is estimated as

$$\mathbb{E}(\hat{Y}|R_m, X_u) \approx \mathcal{X}_u \hat{\beta}$$

After the path probability and the local conditional expectation are computed for each region, the conditional expectation $\mathbb{E}\left(\hat{Y}\big|X_u = x_u^{(i)}\right)$ can be computed for each instance $x^{(i)}$. For SHAP, the value function is computed as

$$c(u) = \mathbb{E}\left(\hat{Y}\big|X_u = x_u^{(i)}\right)$$

For global Shapley, if $N$ is the total number of observational instances and define

$$\bar{m} = \frac{1}{N}\sum_{i=1}^{N} \mathbb{E}\left(\hat{Y}\big|X_u = x_u^{(i)}\right)$$

then the value function for the subset $u$ can be computed as

$$c(u) = Var\left(\mathbb{E}(\hat{Y}|X_u)\right) = \frac{1}{N}\sum_{i=1}^{N}\left(\mathbb{E}\left(\hat{Y}\big|X_u = x_u^{(i)}\right) - \bar{m}\right)^2 \qquad (4)$$

## 3.3 Computational efficiency considerations

Shapley computations require one to compute the value function $c(u)$ for all the possible subsets of features. For $p$ features, there are $2^p$ number of subsets which would be large even for a moderate value of $p$. To improve the computational efficiency, we consider selecting part of all subsets as an approximation.

### 3.3.1 WLS formulation for Shapley

The thresholding method is motivated by the Supplementary result of (Lundberg & Lee, 2017) and has been introduced as an approach by (Aas, et al., 2020). Based on their results, the Shapley values (both the SHAP values of an input and the global Shapley values) can be defined as the optimal solution of the following WLS problem:

$$\min_{\phi_1, \phi_2, \ldots, \phi_p} \sum_{u \subseteq \mathcal{K}} \left(c(u) - \sum_{j \in u} \phi_j\right)^2 w(|u|)$$

where the weight is defined as

$$w(|u|) = \frac{p-1}{\binom{p}{|u|}|u|(p-|u|)}$$

If we rewrite $\phi = (\phi_1, \phi_2, \ldots, \phi_p)^T$, $C = (c(u_1), c(u_2), \ldots, c(u_{2^p}))^T$ with $u_1, u_2, \ldots, u_{2^p}$ are all the enumerations of all $2^p$ subsets of features, $Z$ is a $2^p \times p$ matrix with the $i$-th row of $Z$ is the binary representation of the subset $u_i$, and $B$ is a $2^p \times 2^p$ diagonal matrix with the $i$-th diagonal element is $w(|u_i|)$. Then the WLS problem can be rewritten as

$$\min_{\phi} (C - Z\phi)^T B (C - Z\phi)$$

with its solution is expressed as

$$\phi = (Z^T B Z)^{-1} Z^T B C$$

In the implementation, the infinite weights $w(0) = w(p) = \infty$ can be set to a large constant (e.g., $10^5$) or imposing a constraint into the WLS problem.

The thresholding method uses the WLS formulation to approximate the solution to $\phi$. The thresholding method select part of all subsets $\mathcal{D} \subseteq \{u_1, u_2, \ldots, u_{2^p}\}$ and uses only rows $Z_\mathcal{D}$ of $Z$, $C_\mathcal{D}$ of $C$, and rows and columns $B_\mathcal{D}$ of $B$, and the global Shapley is estimated as

$$\phi \approx (Z_\mathcal{D}^T B_\mathcal{D} Z_\mathcal{D})^{-1} Z_\mathcal{D}^T B_\mathcal{D} C_\mathcal{D} \qquad (5)$$

### 3.3.2 Thresholding method for subset selection

The thresholding method selects important subsets as those that have largest weights in the WLS formulation of Shapley. It relies on two properties of the weight function $w(|u|)$:

1. A subset and its complement have the same weight, i.e., $w(|u|) = w(|\bar{u}|)$ where $\bar{u} = \mathcal{K} \setminus u$;
2. The weight is a decreasing function of the size of a subset or of its complement, whichever is smaller, i.e., $w(|u_1|) < w(|u_2|)$ for any $0 \leq \min(|u_2|, |\bar{u}_2|) < \min(|u_1|, |\bar{u}_1|) \leq p/2$.

Figure 3 provides bar plots for the subsets and their weights in two cases.

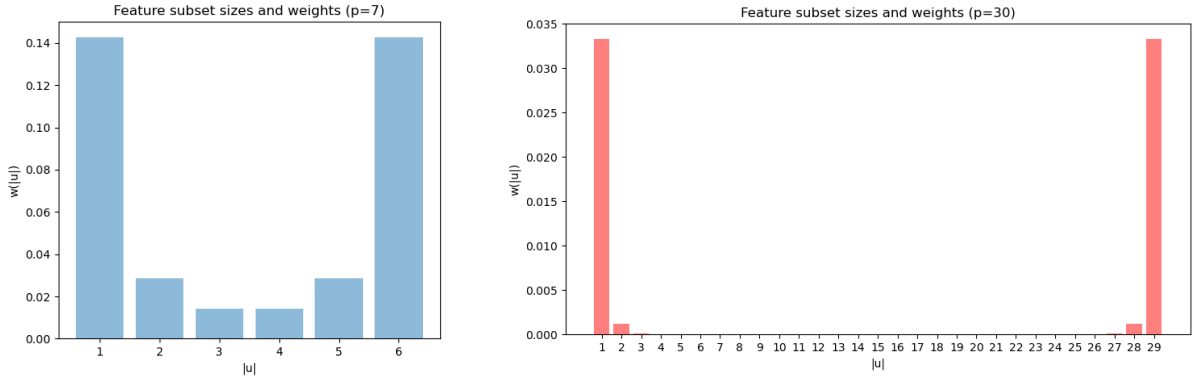

Figure 3: Feature subset sizes and their corresponding weights, when $p = 7$ (left) and $p = 30$ (right). Subsets locating at two tails have significantly larger weights than subsets locating in the middle.

Based on the properties of the weight function, the thresholding method is proposed as follows:

**Definition 2**: Given an integer $\gamma$ satisfying $0 < \gamma \leq \frac{p}{2}$, the *thresholding method* selects the subsets of features as

$$\mathcal{D} = \{u \subseteq \mathcal{K} : |u| \leq \gamma \text{ or } |\bar{u}| \leq \gamma\}$$

Generally, $\gamma$ is a user-provided parameter and its specific value depends on the feature size and the computational resources. For small (and potentially medium) size of features, it is reasonable to choose $\gamma = \left\lfloor \frac{p}{2} \right\rfloor$ ($\left\lfloor \frac{p}{2} \right\rfloor$ denotes the largest integer no greater than $\frac{p}{2}$) that covers all subsets of features. For larger size of features, $\gamma$ is chosen in a way that reflects a trade-off between the accuracy and the computational efficiency.

### 3.4 Detailed algorithm

The detailed algorithm is documented in Table 1.

| | |
|---|---|
| **A unified algorithm for global Shapley and SHAP values computation.** | |
| **Input:** | • A $N \times p$ training data matrix and $N$-dim fitted response vector: $\mathcal{X}, \mathcal{Y}$. ($N$ is size of data points, $p$ is number of features.) <br> • A $N_{exp} \times p$ data matrix and the $N_{exp}$-dim fitted response vector to be explained by SHAP: $\mathcal{X}_{exp}, \mathcal{Y}_{exp}$. (Only required for SHAP computation.) <br> • Number of top important variable: $a$. <br> • Parameter in thresholding method: $\gamma$. |
| **Output:** | (Global or local) Shapley values for input features: $\phi = (\phi_1, \dots, \phi_p)$. |
| 1 | Build a SLIM tree $\mathcal{T}$ using $\mathcal{X}$ and $\mathcal{Y}$. Properly tune the SLIM tree. Denote its $M$ terminal nodes as $R_1, R_2, \dots, R_M$. |
| 2 | For each splitting node in $\mathcal{T}$: <br>   Take the splitting event as a binary response variable, take all other variables (not including the splitting variable) as predictors, and take all data $\mathcal{X}$ that fall into the current node as the training data. Build random forest model and select top $a$ important variables via Gini impurity importance scores, save these variables as set $v$. |
| 3 | For each splitting node in $\mathcal{T}$: |
| 3.1 | Get the top important variable set $v$ of the current splitting node. |
| 3.2 | For each $\emptyset \neq s \subseteq v$: <br>   Take the splitting event as a binary response variable, take all variables in $s$ as predictors, and take all data that fall into the current node as the training data. Build random forest model and save it for later use. |
| 4 | Given $\gamma$, determine the selected subsets of features via thresholding method as $\mathcal{D}$. |
| 5 | For each $u \in \mathcal{D}$: |
| 5.1 | For each terminal node $R_i$: <br>   Find out all the splitting events $A_1, A_2, \dots$ along the path to $R_i$ in a sequential order. <br>   Compute the path probability $p(R_m | X_u = x_u) = \prod_i p(A_i | \cap_{j<i} A_j, X_u = x_u)$ for all $x$ in $\mathcal{X}$. <br>   For each $p(A_i | \cap_{j<i} A_j, X_u)$, there are two cases: <br>   • If $u \cap v \neq \emptyset$, then $p(A_i | \cap_{j<i} A_j, X_u)$ is computed from the model saved in Step 3, with $u \cap v$ as model's predictors. <br>   • If $u \cap v = \emptyset$, then $p(A_i | \cap_{j<i} A_j, X_u)$ is computed empirically from Equation (3). <br>   Compute the local conditional expectation $\mathbb{E}(\hat{Y} | X_u, R_m)$ by fitting a GAM and save it. |
| 5.2 | For both SHAP and global Shapley, for each terminal node $R_i$: <br>   Compute the path probability $p(R_m | X_u = x_u) = \prod_i p(A_i | \cap_{j<i} A_j, X_u = x_u)$ for all $x$ in $\mathcal{X}_{exp}$. <br>   Use the saved GAMs to compute the local conditional expectation $\mathbb{E}(\hat{Y} | X_u, R_m)$ for $\mathcal{X}_{exp}$. <br>   Compute the conditional expectation via $\mathbb{E}(\hat{Y} | X_u) = \sum_{m=1}^{M} p(R_m | X_u) \mathbb{E}(\hat{Y} | X_u, R_m)$ for $\mathcal{X}_{exp}$. |
| 5.3 | For global Shapley, compute the conditional expectation via $\mathbb{E}(\hat{Y} | X_u) = \sum_{m=1}^{M} p(R_m | X_u) \mathbb{E}(\hat{Y} | X_u, R_m)$. Further compute $Var\left(\mathbb{E}(\hat{Y} | X_u)\right)$ via Equation (4). |
| 6 | Compute the global Shapley or SHAP value by aggregating all the value functions via the Shapley WLS formulation in Equation (5). |

*Table 1: Detailed steps of the unified algorithm for global Shapley and SHAP values computation.*

# 4 Simulation Studies

In this section, we describe simulation studies that demonstrate the following three aspects of the proposed method:

1. It provides remarkable improvement in the computation accuracy for global Shapley compared with some methods exist in the current literatures.
2. Both global and local interpretation are unified under the common algorithm, so that the proposed method can offer SHAP values as side results.
3. It offers the way to improve the computational efficiency at the expense of accuracy.

Each of the above aspects is tested by each of the following sub-sections.

## 4.1 Improvement in accuracy

### 4.1.1 Simulation setting

In this sub-section, we numerically demonstrate that, compared with some existing methods in the current literature, the proposed algorithm provides remarkable improvement in the computation accuracy for global Shapley. We compare the performance of the proposed method with some baseline methods from the current literatures, including:

1. *Marginal approximation*, which assumes that features are independent of each other, and the conditional distribution is approximated by the marginal distribution:
$$\mathbb{E}(\hat{f}(X)|X_u) \approx \mathbb{E}_{X_{\bar{u}}}(\hat{f}(x_u, X_{\bar{u}}))$$

2. *Mean Absolute SHAP*, which computes the global Shapley by aggregating Tree SHAP values as
$$\phi_i \approx \frac{1}{n}\sum_{j=1}^{n}|\phi_i^{local}(x^{(j)})|$$
where $\phi_i^{local}(x^{(j)})$ denotes the Tree SHAP value of the $j$-th observation $x$ for feature $i$, and $n$ is the number of observations computed for SHAP.

3. *Empirical Kernel Estimation* by (Aas, et al., 2020), which computes the Gaussian kernel between each observation. Please refer to Appendix 1 for more details.

To conduct the performance comparison, we consider applying different methods over the following simulation scenarios, as shown in Table 2.

| Scenario | p | Data Generating Model |
|---|---|---|
| Linear | 13 | $Y = f(X) = 1.5X_1 + 1.5X_2 + 1.5X_3 + X_4 + 1.4X_5 + 0.5X_6 + 1.8X_7 + 1.8X_8 + 1.6X_{12} + 1.6X_{13}$<br>where $(X_1, X_2, \ldots, X_{13}) \sim \mathcal{N}(0, \Sigma^{(1)})$ |
| Nonlinear | 13 | $Y = f(X) = 1.5X_1 + 1.5\tan(X_2) + 1.5X_3^{\frac{1}{5}} + \ln(X_4) + 1.4X_5 + 0.5X_6^{1/3} + 1.8X_7^3 + 1.8X_8 + 1.6X_{12} + 1.6X_{13}^5$<br>where $(X_1, \tan(X_2), X_3^{\frac{1}{5}}, \ldots, X_{13}^5) \sim \mathcal{N}(0, \Sigma^{(1)})$ |
| Interaction | 6 | $Y = f(X) = X_1 + X_2^2 + X_1 X_2 + X_3 X_5$<br>where $(X_1, X_2, \ldots, X_6) \sim \mathcal{N}(0, \Sigma^{(2)})$ |
| Binary response | 13 | $logit(p) = f(X)$<br>$= 1.5X_1 + 1.5\tan(X_2) + 1.5X_3^{\frac{1}{5}} + \ln(X_4) + 1.4X_5 + 0.5X_6^{1/3} + 1.8X_7^3 + 1.8X_8 + 1.6X_{12} + 1.6X_{13}^5$<br>$Y \sim Bernoulli(p)$<br>where $(X_1, \tan(X_2), X_3^{\frac{1}{5}}, \ldots, X_{13}^5) \sim \mathcal{N}(0, \Sigma^{(1)})$. |

*Table 2: Four different simulation scenarios and their basic settings.*

The structures of the correlation matrices are shown in Figure 4.

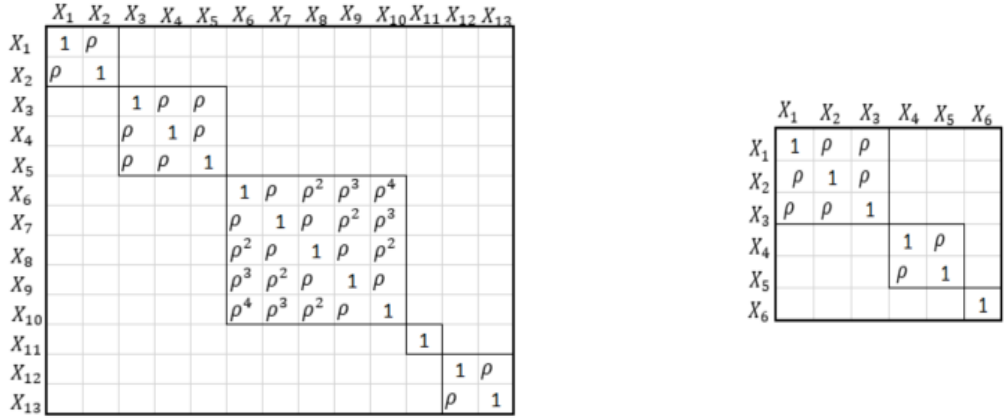

*Figure 4: The correlation structures for $\Sigma^{(1)}$ (left) and $\Sigma^{(2)}$ (right).*

We are particularly interested in testing these four simulation scenarios because of the following reasons. First, it is interesting to test the performance if various effects exist in a model, including linearity, nonlinearity, interaction effect, or binary response variable. Second, input features are designed as group correlated. This helps to check how Shapley attributes importance among group-correlated variables. Finally, for these four scenarios, it is relatively easy to compute the true Shapley values. Generally, the true Shapley values are prohibitively difficult to compute because they are theoretically obtained based on the probability distribution of input features and the true data generation model. To measure the performance of different methods, it is necessary to know the true Shapley values as the benchmark. The true Shapley values are relatively easy to compute for these four scenarios as the conditional distribution of a multivariate normal distribution is easy to compute. In the nonlinear scenario, we assume $(X_1, \tan(X_2), X_3^{\frac{1}{5}}, \ldots, X_{13}^5)$ following the multivariate Gaussian distribution. As the nonlinearity terms are monotone functions of input features coming from the linear scenario, it is easy to derive that the true global Shapley values of the non-linear scenario will be the same as those of the linear scenario.

In the simulation, we examine the performance under different values for the correlation $\rho$, including $0, 0.2, 0.4, 0.5, 0.75, 0.9, 0.99$. For each simulation scenario, we compute the true global Shapley values $\phi$. For each method, we compute the estimate global Shapley values $\hat{\phi}$, and measure the performance by the error metric

$$error = \frac{||\phi - \hat{\phi}||_F^2}{||\phi||_F^2} = \frac{\sum_{i=1}^p (\phi_i - \hat{\phi}_i)^2}{\sum_{i=1}^p \phi_i^2}$$

where $\phi$ and $\hat{\phi}$ are expressed in percentage. Given the value of the correlation $\rho$, we generate 20,000 samples (i.e., the size of $\mathcal{X}$ and $\mathcal{Y}$) for $X$ and $Y$ as the dataset. Additionally, we generate additional 2,000 samples to test the out-of-sample performance of SLIM tree (see the results in Appendix 2). The dataset is used to compute the global Shapley values while the test dataset is used to test the performance of the SLIM tree. The performance results are shown in Appendix 2. For the linear, non-linear, and interaction simulation scenarios, the generated data $X$ and $Y$ are directly used to compute the global Shapley values. This means that we assume the machine learning model is the same as the true model (i.e., $\hat{f}(X) = f(X)$). Such a simplicity facilitates the performance comparison between different methods and removes the computation errors that are caused by $\hat{f}(X)$ not able to perfectly fit the response surface. For the binary response case, the generated data are used to build an Xgboost model, which computes the classification probability $\hat{p}$ for each observation. Then the global Shapley is computed over the log-odds space, i.e.,

$$\hat{f}(X) = \hat{\eta} = logit(\hat{p}) = \ln\left(\frac{\hat{p}}{1 - \hat{p}}\right)$$

More set-up details can be found in Appendix 2.

### 4.1.2 Results and findings

All the results are produced using Wells Fargo Corporate Model Risk (CMoR) computing servers. For each simulation scenario, the results for the correlation $\rho = 0.9, 0.4, 0$ are shown in the following tables. Full results are shown in Appendix 3. In the tables, "Oracle" refers to the true global Shapley values (the benchmark), "MBT" means the model-based tree method, "Marginal" means using the marginal approximation, "Empirical" represents the empirical conditional estimation, and "M.A.SHAP" refers to the mean absolute aggregation of Tree SHAP values. All results are presented in Table 3, Table 4, Table 5, and Table 6. Three findings from these tables are:

1. Results from the proposed method (i.e., "MBT" in the tables) have small computational error relative to the oracle results. The individual Shapley value is close to the oracle value, and it is easy to check that the feature importance ranks are also similar.
2. For a certain scenario, although the model formulation is unchanged, the oracle Shapley values will change if the feature correlations are changed. This indicates that global Shapley based on the value function in Equation (1) not only reflects the behavior of a model, but also the mechanism of the underlying data.
3. If a feature is not a variable in the true data generating model and it is independent of other variables (such as $X_{11}$ in the linear scenario and $X_6$ in the interaction scenario), then its global Shapley value is zero. This property is preserved in the results of the proposed method. On the other hand, marginal approximation method attributes zero contribution to the features not used in the model $\hat{f}(X)$, such as features $X_9, X_{10}, X_{11}$ in the linear, nonlinear, and binary response scenarios, and features $X_4, X_6$ in the interaction scenario. The proposed algorithm attributes nonzero scores to these features if they are correlated to at least one feature that is used in the model $\hat{f}(X)$. Such a difference is consistent with the argument about using marginal or conditional distribution in Shapley (Chen, et al., 2020). That is, using conditional distribution will attribute importance scores to correlated features not used in the model.

| | | | | | | | Linear Scenario | | | | | | | | |
|---|---|---|---|---|---|---|---|---|---|---|---|---|---|---|---|
| $\rho$ | | X1 | X2 | X3 | X4 | X5 | X6 | X7 | X8 | X9 | X10 | X11 | X12 | X13 | Error |
| 0.9 | Oracle | 8.88 | 8.88 | 10.07 | 9.50 | 9.95 | 6.74 | 8.47 | 8.09 | 5.47 | 3.74 | 0.00 | 10.11 | 10.11 | - |
| | MBT | 9.14 | 9.03 | 9.92 | 9.36 | 9.82 | 6.67 | 8.51 | 8.04 | 5.49 | 3.87 | 0.01 | 10.06 | 10.08 | 0.00020 |
| | Marginal | 9.11 | 8.86 | 11.19 | 7.39 | 10.49 | 3.66 | 14.71 | 14.57 | 0.00 | 0.00 | 0.00 | 10.01 | 10.02 | 0.15966 |
| | Empirical | 2.94 | 10.81 | 7.75 | 11.58 | 5.75 | 10.50 | 7.08 | 8.55 | 14.95 | 4.13 | 7.14 | 4.05 | 4.78 | 0.32842 |
| | M.A.SHAP | 11.68 | 9.57 | 11.84 | 6.21 | 9.30 | 2.57 | 12.85 | 13.26 | 0.09 | 0.04 | 0.02 | 12.06 | 10.51 | 0.15080 |
| 0.4 | Oracle | 9.54 | 9.54 | 10.75 | 7.19 | 9.97 | 3.22 | 13.89 | 12.91 | 1.17 | 0.11 | 0.00 | 10.86 | 10.86 | - |
| | MBT | 9.68 | 9.39 | 10.48 | 7.02 | 9.87 | 3.17 | 14.18 | 12.86 | 1.14 | 0.07 | 0.00 | 11.04 | 11.10 | 0.00032 |
| | Marginal | 9.94 | 9.62 | 10.84 | 6.54 | 10.21 | 2.26 | 14.72 | 13.94 | 0.00 | 0.00 | 0.00 | 11.08 | 10.84 | 0.00486 |
| | Empirical | 4.31 | 7.24 | 7.67 | 12.08 | 6.76 | 13.23 | 5.70 | 6.50 | 7.38 | 3.82 | 8.61 | 10.19 | 6.52 | 0.40764 |
| | M.A.SHAP | 10.49 | 10.34 | 10.73 | 7.11 | 9.79 | 3.25 | 12.93 | 12.59 | 0.06 | 0.07 | 0.03 | 11.20 | 11.42 | 0.00404 |
| 0 | Oracle | 10.44 | 10.44 | 10.44 | 4.64 | 9.09 | 1.16 | 15.03 | 15.03 | 0.00 | 0.00 | 0.00 | 11.87 | 11.87 | - |
| | MBT | 10.38 | 10.08 | 10.34 | 4.90 | 8.86 | 1.12 | 14.73 | 15.26 | 0.00 | 0.00 | 0.00 | 11.84 | 12.50 | 0.00068 |
| | Marginal | 11.18 | 10.05 | 10.46 | 4.73 | 8.95 | 1.09 | 14.51 | 15.13 | 0.00 | 0.00 | 0.00 | 11.70 | 12.20 | 0.00098 |
| | Empirical | 11.15 | 6.90 | 5.26 | 4.57 | 17.80 | 2.42 | 6.36 | 17.77 | 4.98 | 5.86 | 8.33 | 4.59 | 4.02 | 0.38012 |
| | M.A.SHAP | 10.52 | 10.77 | 10.68 | 5.52 | 9.64 | 0.01 | 13.96 | 14.18 | 0.00 | 0.00 | 0.00 | 12.28 | 12.44 | 0.00422 |

*Table 3: Results for linear scenario, with $\rho = 0.9, 0.4, 0$. Please refer to Appendix 3 for the full results.*

| Non-linear Scenario | | | | | | | | | | | | | | | |
|---|---|---|---|---|---|---|---|---|---|---|---|---|---|---|---|
| $\rho$ | | X1 | X2 | X3 | X4 | X5 | X6 | X7 | X8 | X9 | X10 | X11 | X12 | X13 | Error |
| 0.9 | Oracle | 8.88 | 8.88 | 10.07 | 9.50 | 9.95 | 6.74 | 8.47 | 8.09 | 5.47 | 3.74 | 0.00 | 10.11 | 10.11 | - |
| | MBT | 9.14 | 9.02 | 9.93 | 9.37 | 9.82 | 6.68 | 8.51 | 8.04 | 5.49 | 3.87 | 0.01 | 10.06 | 10.07 | 0.00019 |
| | Marginal | 9.11 | 8.86 | 11.19 | 7.39 | 10.49 | 3.66 | 14.71 | 14.57 | 0.00 | 0.00 | 0.00 | 10.01 | 10.02 | 0.15966 |
| | Empirical | 3.10 | 8.86 | 4.13 | 10.78 | 7.29 | 11.07 | 11.47 | 6.56 | 15.57 | 3.59 | 5.88 | 4.02 | 7.66 | 0.32651 |
| | M.A.SHAP | 11.86 | 9.31 | 11.91 | 6.14 | 9.22 | 2.54 | 12.81 | 13.42 | 0.11 | 0.07 | 0.08 | 12.13 | 10.41 | 0.15407 |
| 0.4 | Oracle | 9.54 | 9.54 | 10.75 | 7.19 | 9.97 | 3.22 | 13.89 | 12.91 | 1.17 | 0.11 | 0.00 | 10.86 | 10.86 | - |
| | MBT | 9.69 | 9.29 | 10.49 | 7.01 | 9.87 | 3.18 | 14.17 | 12.92 | 1.13 | 0.07 | 0.00 | 11.06 | 11.10 | 0.00036 |
| | Marginal | 9.94 | 9.62 | 10.84 | 6.54 | 10.21 | 2.26 | 14.72 | 13.94 | 0.00 | 0.00 | 0.00 | 11.08 | 10.84 | 0.00097 |
| | Empirical | 3.97 | 6.79 | 5.72 | 15.69 | 6.04 | 14.58 | 7.28 | 4.86 | 5.47 | 3.08 | 6.92 | 8.83 | 10.75 | 0.44377 |
| | M.A.SHAP | 10.56 | 10.33 | 10.74 | 7.09 | 9.76 | 3.24 | 12.89 | 12.65 | 0.05 | 0.03 | 0.03 | 11.16 | 11.47 | 0.00428 |
| 0 | Oracle | 10.44 | 10.44 | 10.44 | 4.64 | 9.09 | 1.16 | 15.03 | 15.03 | 0.00 | 0.00 | 0.00 | 11.87 | 11.87 | - |
| | MBT | 10.37 | 10.07 | 10.35 | 4.89 | 8.86 | 1.13 | 14.67 | 15.33 | 0.00 | 0.00 | 0.00 | 11.84 | 12.49 | 0.00073 |
| | Marginal | 11.18 | 10.05 | 10.46 | 4.73 | 8.95 | 1.09 | 14.51 | 15.13 | 0.00 | 0.00 | 0.00 | 11.70 | 12.20 | 0.00098 |
| | Empirical | 9.63 | 7.67 | 4.54 | 4.51 | 14.62 | 6.99 | 8.93 | 17.00 | 5.45 | 4.20 | 8.20 | 4.51 | 3.75 | 0.32890 |
| | M.A.SHAP | 10.52 | 10.62 | 10.61 | 6.34 | 9.84 | 2.33 | 13.18 | 13.22 | 0.00 | 0.00 | 0.00 | 11.60 | 11.74 | 0.00999 |

Table 4: Results for non-linear scenario, with $\rho = 0.9, 0.4, 0$. Please refer to Appendix 3 for the full results.

| Interaction Scenario | | | | | | | | |
|---|---|---|---|---|---|---|---|---|
| | | X1 | X2 | X3 | X4 | X5 | X6 | Error |
| 0.9 | Oracle | 29.09 | 40.30 | 24.43 | 2.38 | 3.80 | 0.00 | - |
| | MBT | 28.07 | 38.88 | 23.74 | 2.82 | 5.64 | 0.84 | 0.00253 |
| | Marginal | 29.69 | 60.72 | 4.96 | 0.00 | 4.64 | 0.00 | 0.25986 |
| | Empirical | 20.71 | 37.40 | 9.63 | 8.44 | 12.30 | 11.52 | 0.17470 |
| | M.A.SHAP | 29.97 | 52.30 | 5.87 | 0.32 | 11.48 | 0.06 | 0.17895 |
| 0.4 | Oracle | 23.76 | 59.35 | 9.21 | 0.45 | 7.23 | 0.00 | - |
| | MBT | 24.74 | 57.09 | 9.44 | 0.98 | 7.05 | 0.71 | 0.00164 |
| | Marginal | 29.12 | 57.07 | 6.97 | 0.00 | 6.84 | 0.00 | 0.00929 |
| | Empirical | 6.66 | 49.58 | 11.06 | 9.36 | 10.34 | 12.99 | 0.15365 |
| | M.A.SHAP | 30.18 | 47.00 | 10.98 | 0.33 | 11.43 | 0.08 | 0.05075 |
| 0 | Oracle | 30.00 | 50.00 | 10.00 | 0.00 | 10.00 | 0.00 | - |
| | MBT | 30.43 | 47.29 | 10.12 | 1.07 | 10.06 | 1.03 | 0.00270 |
| | Marginal | 31.01 | 48.65 | 10.74 | 0.00 | 9.60 | 0.00 | 0.00098 |
| | Empirical | 19.13 | 20.82 | 17.05 | 16.50 | 13.19 | 13.33 | 0.41099 |
| | M.A.SHAP | 32.89 | 41.21 | 12.86 | 0.07 | 12.88 | 0.09 | 0.02838 |

Table 5: Results for interaction scenario, with $\rho = 0.9, 0.4, 0$. Please refer to Appendix 3 for the full results.

| | | \multicolumn{14}{c}{Binary Response Scenario} |
|---|---|---|---|---|---|---|---|---|---|---|---|---|---|---|
| $\rho$ | | X1 | X2 | X3 | X4 | X5 | X6 | X7 | X8 | X9 | X10 | X11 | X12 | X13 | Error |
| 0.9 | Oracle | 8.88 | 8.88 | 10.07 | 9.50 | 9.95 | 6.74 | 8.47 | 8.09 | 5.47 | 3.74 | 0.00 | 10.11 | 10.11 | - |
| | MBT | 9.33 | 9.39 | 9.81 | 9.06 | 9.63 | 6.49 | 8.18 | 8.04 | 5.38 | 3.86 | 0.12 | 10.27 | 10.44 | 0.00131 |
| | Marginal | 9.11 | 8.86 | 11.19 | 7.39 | 10.49 | 3.66 | 14.71 | 14.57 | 0.00 | 0.00 | 0.00 | 10.01 | 10.02 | 0.15966 |
| | Empirical | 3.30 | 8.88 | 4.23 | 9.20 | 6.98 | 9.93 | 12.58 | 7.35 | 17.24 | 3.35 | 5.78 | 3.78 | 7.40 | 0.36565 |
| | M.A.SHAP | 10.52 | 9.14 | 9.87 | 6.45 | 11.07 | 1.97 | 14.34 | 13.08 | 0.99 | 0.40 | 0.51 | 10.60 | 11.07 | 0.14572 |
| 0.4 | Oracle | 9.54 | 9.54 | 10.75 | 7.19 | 9.97 | 3.22 | 13.89 | 12.91 | 1.17 | 0.11 | 0.00 | 10.86 | 10.86 | - |
| | MBT | 9.14 | 9.68 | 9.96 | 7.20 | 9.56 | 3.37 | 14.07 | 12.59 | 1.48 | 0.19 | 0.14 | 11.05 | 11.57 | 0.00170 |
| | Marginal | 9.86 | 9.28 | 10.92 | 6.58 | 10.03 | 2.28 | 14.85 | 14.06 | 0.01 | 0.01 | 0.01 | 11.13 | 10.96 | 0.00486 |
| | Empirical | 4.04 | 6.60 | 5.39 | 17.17 | 5.91 | 14.74 | 7.05 | 5.24 | 5.18 | 3.37 | 5.75 | 9.24 | 10.32 | 0.45909 |
| | M.A.SHAP | 10.09 | 10.16 | 10.94 | 6.46 | 9.95 | 3.80 | 13.22 | 12.06 | 0.75 | 0.54 | 0.50 | 10.91 | 10.62 | 0.00323 |
| 0 | Oracle | 10.44 | 10.44 | 10.44 | 4.64 | 9.09 | 1.16 | 15.03 | 15.03 | 0.00 | 0.00 | 0.00 | 11.87 | 11.87 | - |
| | MBT | 10.36 | 10.75 | 9.59 | 4.42 | 8.83 | 1.22 | 14.68 | 15.01 | 0.13 | 0.14 | 0.14 | 12.35 | 12.37 | 0.00136 |
| | Marginal | 11.18 | 10.05 | 10.46 | 4.73 | 8.95 | 1.09 | 14.51 | 15.13 | 0.00 | 0.00 | 0.00 | 11.70 | 12.20 | 0.00098 |
| | Empirical | 9.83 | 8.03 | 4.17 | 4.27 | 16.00 | 7.74 | 7.85 | 16.35 | 5.41 | 4.34 | 9.02 | 4.31 | 2.67 | 0.39564 |
| | M.A.SHAP | 10.20 | 10.54 | 9.83 | 7.35 | 9.68 | 3.76 | 12.89 | 12.36 | 0.32 | 0.43 | 0.37 | 11.05 | 11.21 | 0.02413 |

*Table 6: Results for binary response scenario, with $\rho = 0.9, 0.4, 0$. Please refer to Appendix 3 for the full results.*

## 4.2 SHAP values as side results of global Shapley

### 4.2.1 Simulation setting

The proposed method can offer SHAP values for each observation as side results. As an illustration, we examine the interaction scenario, i.e.,

$$Y = f(X) = X_1 + X_2^2 + X_1 X_2 + X_3 X_5$$

where $(X_1, X_2, \ldots, X_6) \sim \mathcal{N}(0, \Sigma^{(2)})$ and the structure of $\Sigma^{(2)}$ is shown in Figure 4. This simulation scenario contains the main effect, nonlinearity, and interaction. We consider the correlation $\rho = 0.99, 0.8, 0.5, 0.3, 0$. We generate 20,000 samples (i.e., the size of $\mathcal{X}$ and $\mathcal{Y}$) for the dataset to build the SLIM tree. We take the first 1,000 observations as the data points to be explained (i.e., the size of $\mathcal{X}_{exp}$ and $\mathcal{Y}_{exp}$ is 1,000) and compare the SHAP values with the theoretical SHAP values (which are the SHAP values that are computed based on the distribution of the input features and the true data generating model). To compare the performance, we choose the Tree SHAP as the baseline method. In the Tree SHAP method, we consider building three machine learning models as comparison:

1. Random forest (RF)
2. Xgboost tree model with all features are sampled in each tree (xgb-f)

3. Xgboost tree model with partial features are sampled in each tree (xgb-p). In our simulation, we choose the feature subsample ratio as 0.6 when constructing each tree. Subsampling of features occurs once for every tree constructed.

All the machine learning models are properly tuned. For each method and each feature, both the true and the estimated SHAP values are plotted in a scatter plot. A close alignment of scatter points along a diagonal line is an indication of a good performance.

We are interested in comparing xgb-f and xgb-p because the two models may produce different feature importance scores. On the one hand, in xgb-f model, each tree is fitted using all the available features. Therefore, if two features are highly correlated and only one feature is used in the fitting, then another feature is not used in the tree and would be assigned zero importance score. On the other hand, in xgb-p model, each tree is fitted using features sampled from all the input features. Thus, if two features are highly correlated and as long as one of the features is selected in the subsampling, then the feature would be assigned non-zero importance score.

### 4.2.2 Results and conclusions

We consider the performance under different choice of the correlation $\rho = 0.99, 0.8, 0.5, 0.3, 0$. The true and estimated SHAP values are plotted in Figure 5 - Figure 9. From these figures, we can see that when the correlation is non-zero, the scatter points of SHAP values are generally more aligned along the diagonal lines for the proposed algorithm than for the RF, xgb-f, and xgb-p methods. Therefore, the proposed algorithm provides SHAP values that are closer to the true values than other methods. In addition, we observe that when the correlation is zero, scatter points are closer to the diagonal lines for all the four methods, while when the correlation becomes larger, the points become scatter for the RF, xgb-p, and xgb-f. We believe this is an indication that the proposed algorithm is better to capture the correlations among features in the SHAP values.

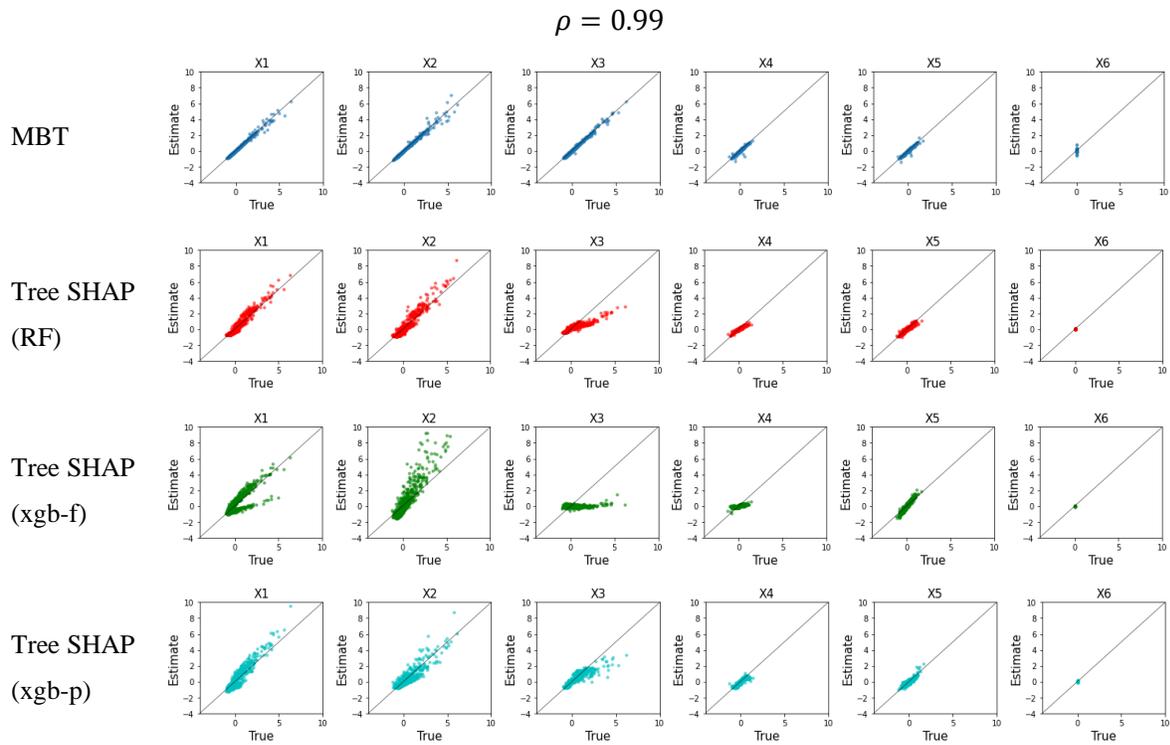

*Figure 5: SHAP values for 1,000 predictions using different methods for $\rho = 0.99$. Y-axis represents the estimated values; X-axis represents the true values.*

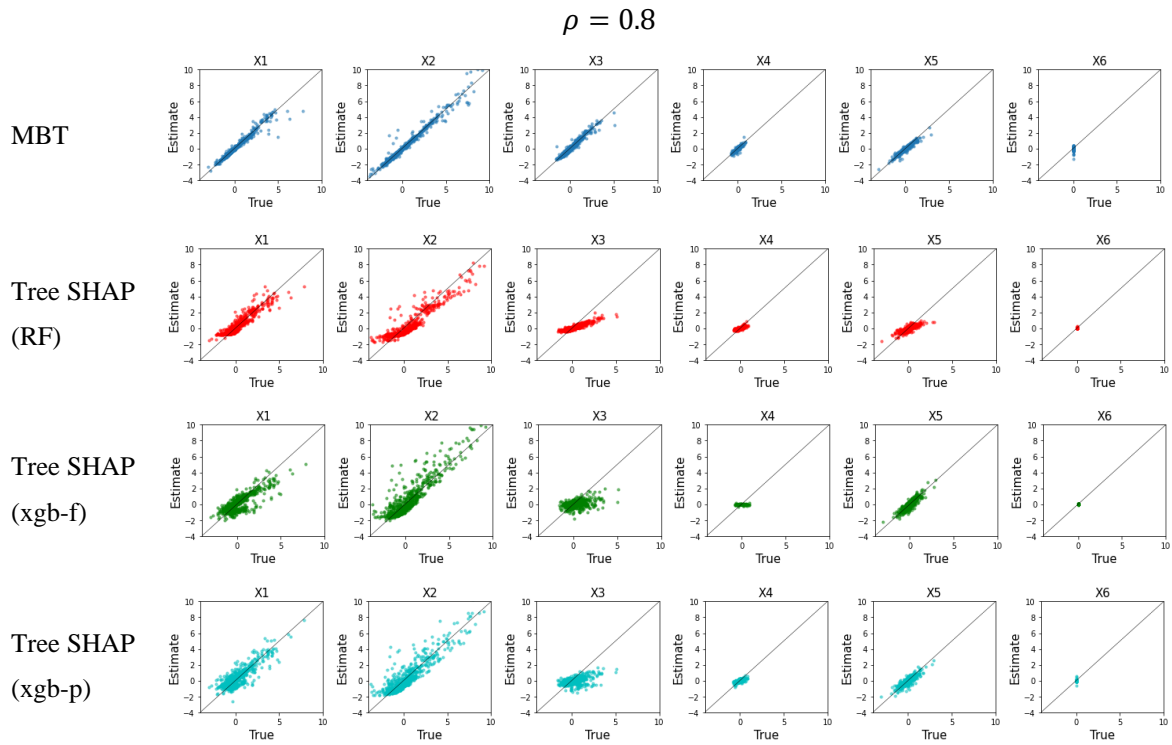

*Figure 6: SHAP values for 1,000 predictions using different methods for $\rho = 0.8$. Y-axis represents the estimated values; X-axis represents the true values.*

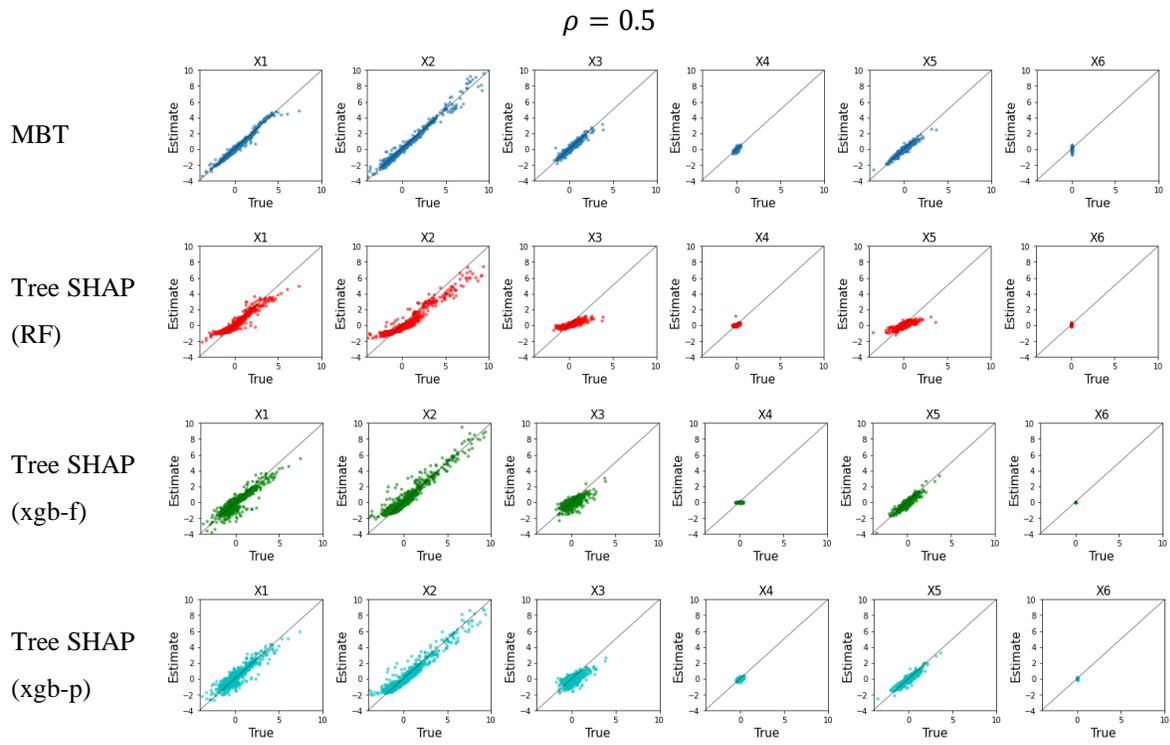

*Figure 7: SHAP values for 1,000 predictions using different methods for $\rho = 0.5$. Y-axis represents the estimated values; X-axis represents the true values.*

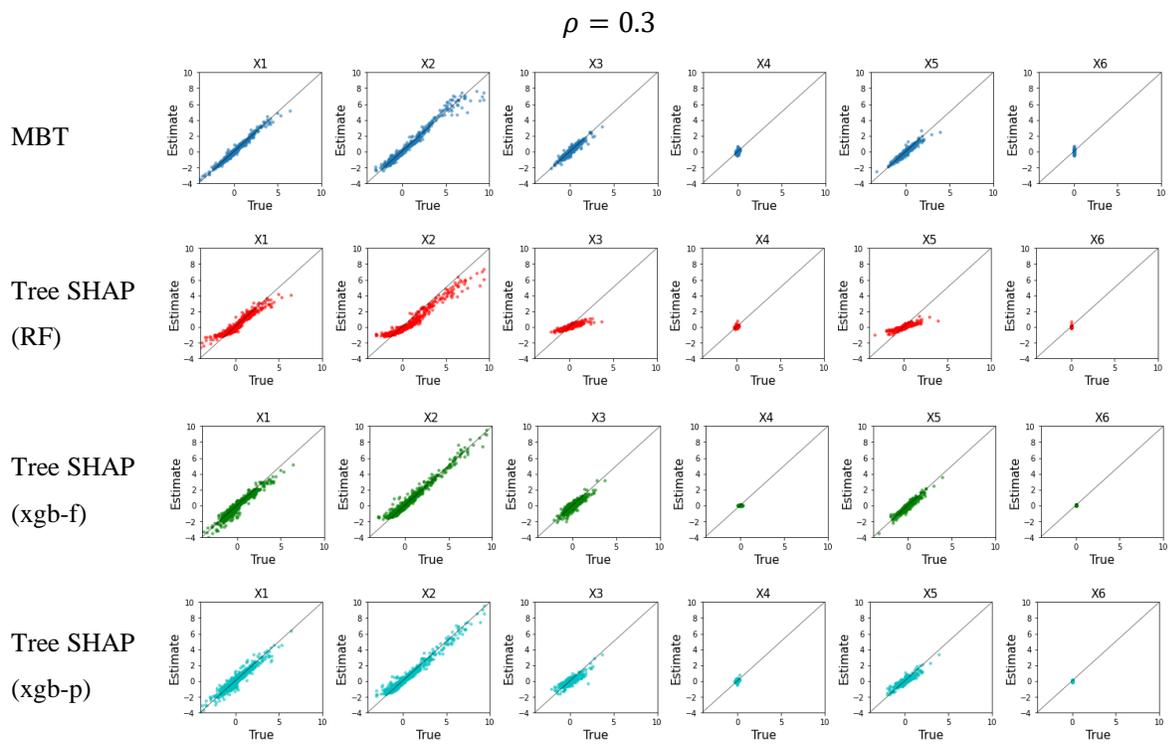

*Figure 8: SHAP values for 1,000 predictions using different methods for $\rho = 0.3$. Y-axis represents the estimated values; X-axis represents the true values.*

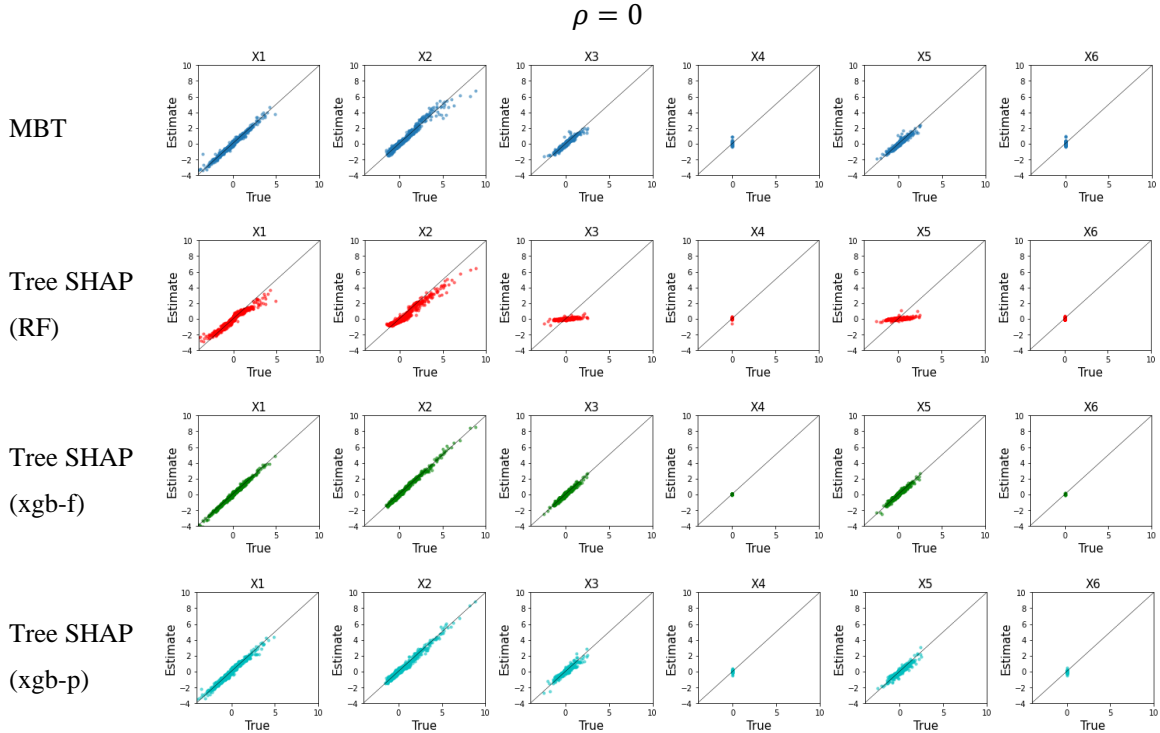

*Figure 9: SHAP values for 1,000 predictions using different methods for $\rho = 0$. Y-axis represents the estimated values; X-axis represents the true values.*

## 4.3 Computational efficiency

The proposed method offers a way to improve the computational efficiency at the expense of accuracy. For illustration purpose, we consider the linear, nonlinear, and interaction scenarios and examine the running time as well as the computational accuracy. Specifically, for linear and nonlinear cases, $p = 13$ and we examine for subset with sizes $\gamma \leq 1, \gamma \leq 2, \gamma \leq 3, \gamma \leq 4$, and with all possible sizes. For interaction case, $p = 6$, and we examine for $\gamma \leq 1, \gamma \leq 2$, and all possible sizes. Figure 10 shows the running time and computational accuracy for each case[4].

### 4.3.1 Conclusions

From the results, it is easy to see that the computation efficiency can be improved by choosing only part of the feature subsets via a smaller $\gamma$. Meanwhile, the computation efficiency is improved at the expense of reducing the accuracy. Therefore, the choice of $\gamma$ reflects a trade-off between the requirements of the computation accuracy and efficiency. In addition, we notice that for all the simulation scenarios, at a certain value of $\gamma$, the computational error is greater when the features are more strongly correlated. Such a result may indicate that, in general, when input features are strongly correlated, a larger value of $\gamma$ is required, while when they are weakly correlated, a smaller value of $\gamma$ is more acceptable. In our simulation, when features are independent (i.e., $\rho = 0$), using $\gamma = 1$ offers the same accurate results as using all possible feature subsets.

---

[4] The running time heavily depends on the computing platform. The results are generated using Wells Fargo CMoR computing platform with parallel computing.

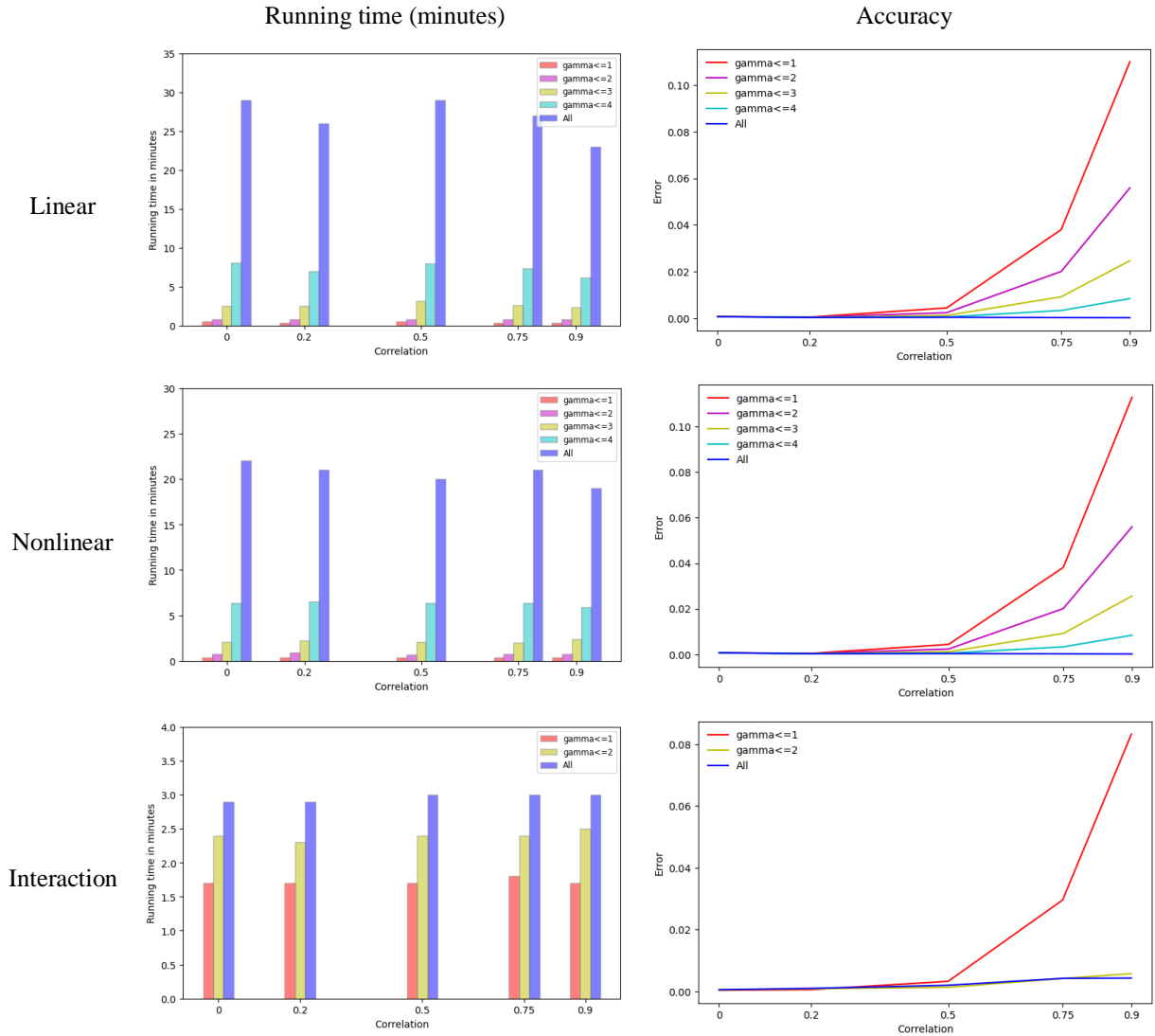

*Figure 10: Running time and accuracy comparison.*

## 4.4 Bike sharing case study

### 4.4.1 Dataset description

The dataset[5] contains 17,379 records. The dataset contains 11 predictor variables and 1 response variable.

| Variable | Description |
|---|---|
| **Predictor variables** | |
| season | 1: spring, 2: summer, 3: fall, 4: winter. |
| mnth | The months from 1 to 12. |
| hr | The hours of a day, from 0 to 23. |
| holiday | Whether the day is a holiday (1) or not (0). |
| weekday | Day of the week from 0 to 6. |
| workingday | If the day is neither in weekend nor is a holiday, it is 1, otherwise, it is 0. |
| weathersit | The weather condition. 1 for clear, few clouds. 2 for Mist and cloudy, 3 for light snow or rain or thunderstorm, 4 for heavy rain, snow, thunderstorm. |
| temp | Normalized temperature in Celsius. |
| atemp | Normalized feeling temperature in Celsius. |
| hum | Normalized humidity. |
| windspeed | Normalized wind speed. |
| **Response variable** | |
| cnt | The count of total rental bikes including both casual and registered users. |

*Table 7: Variables description of the bike sharing dataset.*

The correlation heatmap of all variables is shown in Figure 11.

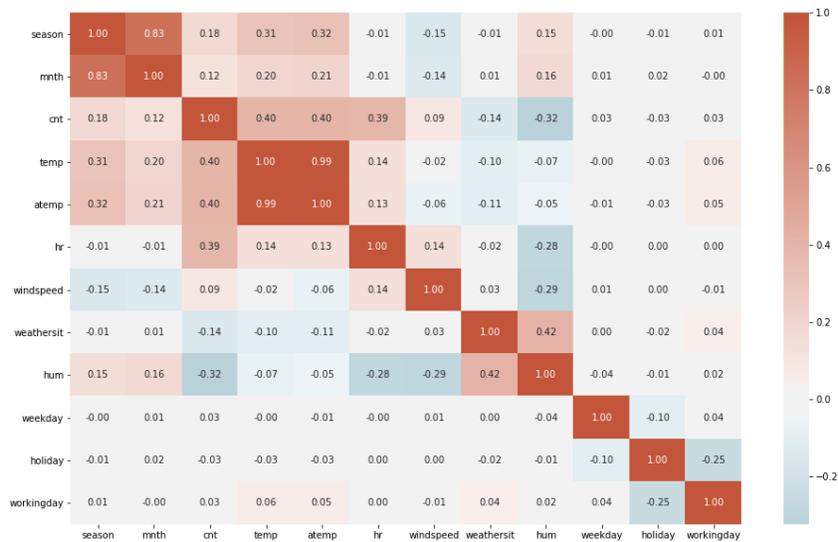

*Figure 11: The Spearman correlation heatmap of all variables in bike sharing dataset.*

---
[5] The bike sharing dataset is publicly available in http://capitalbikeshare.com/system-data.

### 4.4.2 Results and analysis

We used the following steps to compute the global Shapley and SHAP values for this dataset.

1. **Data preprocessing.** We take the log for the response variable and use the log-counts as the response in the model. In addition, we notice that the feature "temp" is strongly correlated with "atemp", and the feature "mnth" is potentially a more refined feature of the feature "season". Therefore, we remove "atemp" and "season" from the list of predictor variables. We treat all the remaining selected predictor variables numerically in the model fitting.

2. **Split the datasets and fitting a machine learning model**. We split the dataset into the training set (size 11,735) and test set (size 5,644). The training data are used to fit an Xgboost regression model while the test data are used to test the fitting performance of the model. The Xgboost model is tuned properly. The model's performance is measured by the mean squared error (MSE) and the $R^2$ score, and the results are shown in Table 8.

|  | Training data | Test data |
|---|---|---|
| MSE | 0.109 | 0.155 |
| $R^2$ | 0.951 | 0.928 |

*Table 8: Performance of the fitted Xgboost regression model.*

3. **Compute the global Shapley of the fitted model.** Use the training dataset and the training fitted response to compute the global Shapley of the model. In the global Shapley computation, the training data and their corresponding fitted response from the Xgboost model are used to fit and tune a SLIM tree. The SLIM tree is visualized in Figure 12.

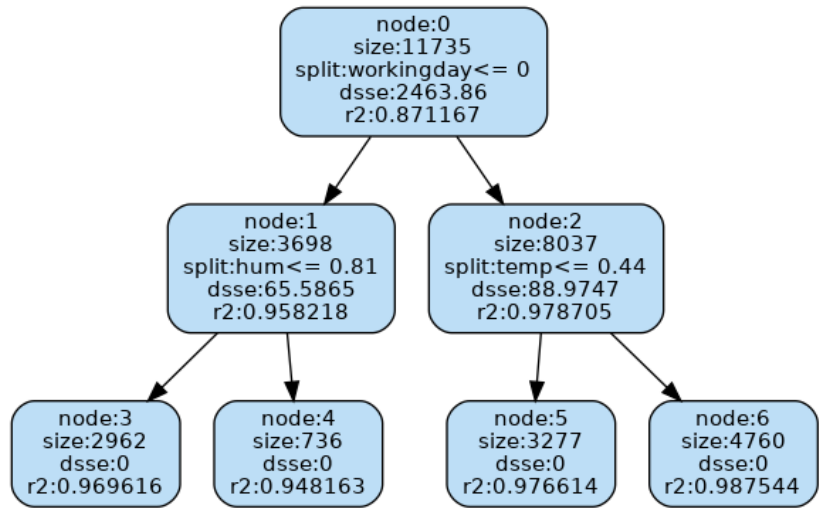

*Figure 12: The fitted SLIM tree in the computation of global Shapley.*

The test data set and its corresponding response (both the fitted and the original) are used to measure the performance of the SLIM tree, which is shown in Table 9. Table 9 also shows the fidelity and accuracy on the model performance. According to (Hu, et al., 2022 (to appear)), in our context, fidelity means how well the surrogate model reproduced the original XGBoost predictions and accuracy means how well SLIM and XGBoost predicted their original responses. The fidelity values show that SLIM is very close to the original XGBoost model with $R^2$ of 98%. The accuracy values indicate that the gap between the test data and the training data is smaller for SLIM, suggesting that there is less overfitting with SLIM.

|  | Performance against fitted response (SLIM-Fidelity) | | Performance against original response (SLIM-Accuracy) | | Performance of Xgboost model (ML-Accuracy) | |
|---|---|---|---|---|---|---|
|  | Train | Test | Train | Test | Train | Test |
| MSE | 0.042 | 0.036 | 0.182 | 0.191 | 0.109 | 0.155 |
| $R^2$ | 0.980 | 0.982 | 0.918 | 0.912 | 0.951 | 0.928 |

*Table 9: Performance of the SLIM tree in the global Shapley computation, compared with the performance of the ML model.*

In the global Shapley computation, we select all the possible feature subsets (i.e., $\gamma = 4$). The global Shapley results (in percentage) are shown in Figure 13. We compare the results from the following methods:

1) The permutation-based importance based on the Xgboost model.
2) The proposed algorithm (MBT).
3) The marginal approximation method. This method assumes that features are independent of each other, thus the conditional distribution in the value function is computed as the marginal distribution, and the fitted response is computed from the Xgboost model in Step 1.
4) The SHAP values aggregation (M.A.SHAP). In the M.A.SHAP, we aggregate the Tree SHAP values of all the predictions of all the training dataset computed from the Xgboost model in Step 1.

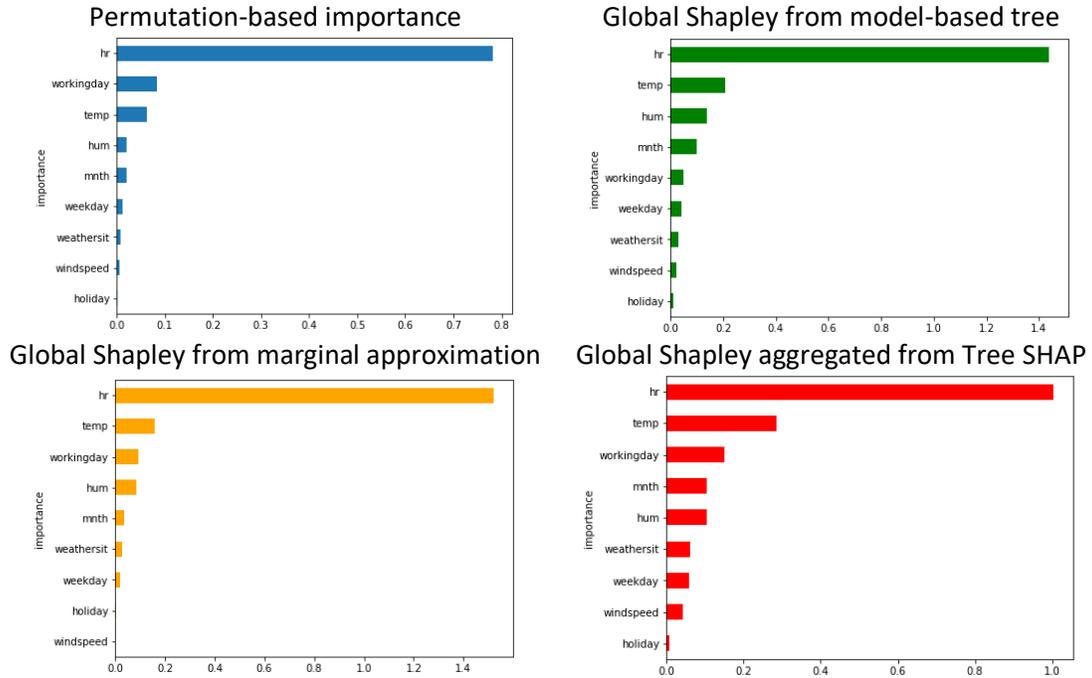
*Figure 13: Importance plot from each method.*

Below are the explanations we provide for the results:
1) All methods agree that the feature "hr" is the most important feature in the dataset.
2) Rankings of features "temp", "hum", "mnth", "weekday", "weathersit", "windspeed", and "holiday" are relatively consistent for all the methods, as the rankings of these features are within 2-level difference.

We plot the feature important scores for every feature in all terminal nodes in Figure 14. As the figure illustrates, the importance of "workingday" in each region is zero. This indicates that the effect of "workingday" is limited to the computation of path probability and has little impact on the computation of local conditional expectations. Thus, the overall importance of "workingday" decreases.

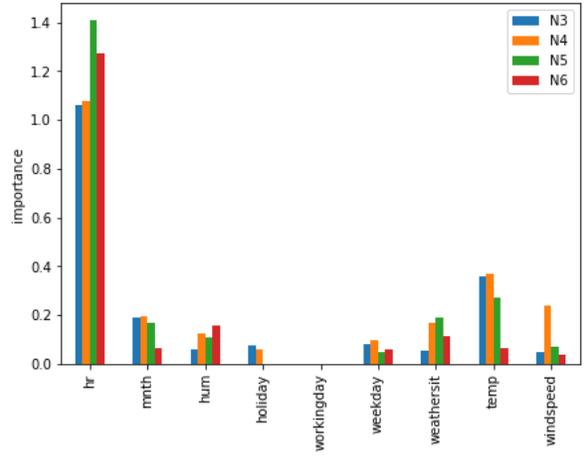

*Figure 14: The importance of each variable as modeled by the model in each terminal node of the SLIM tree.*

4. **Compute the SHAP values for all the training data.** We are also interested in comparing the SHAP values from the proposed method (MBT) and the Tree SHAP method. We compute the SHAP values for all the training data points. In Tree SHAP, the Xgboost model fitted in Step 2 is used as the model to be explained. All the scatter plots of SHAP values for each feature are presented in Figure 15Figure 15.

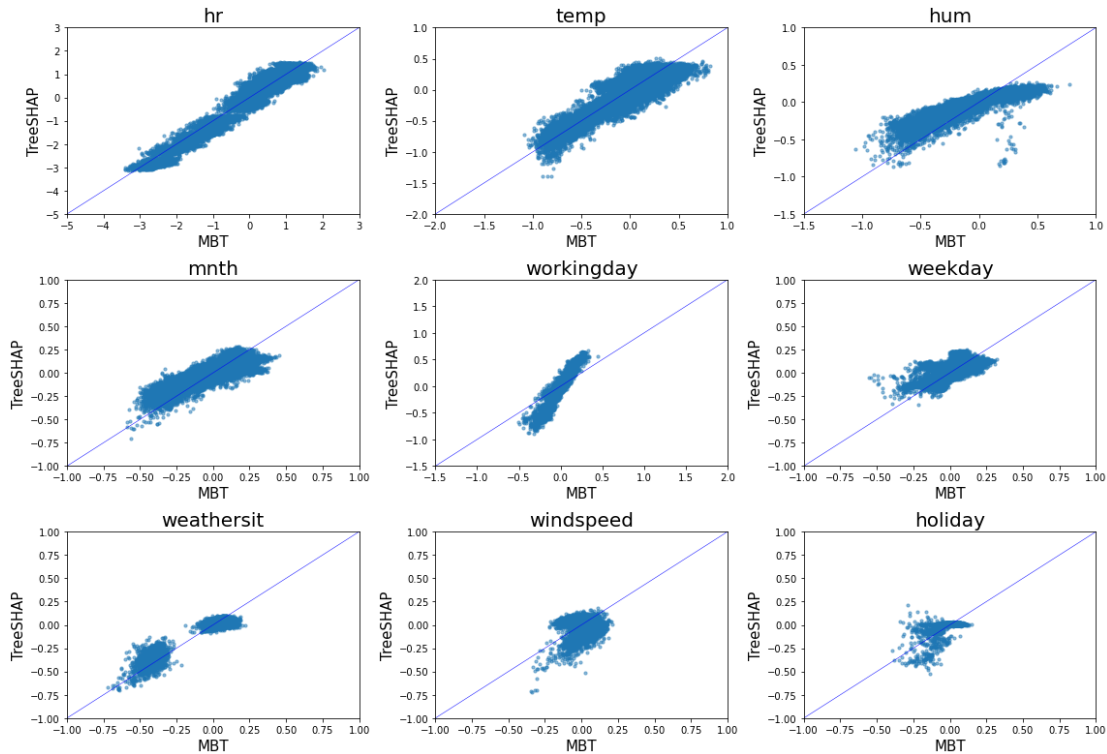

*Figure 15: Scatter plots of SHAP values of Tree SHAP against the proposed method.*

Two conclusions can be made from the scatter plots. First, in the plots of the feature "hr", "temp", "hum", "mnth", and "workingday", it is obvious that points are scattered along a diagonal line. While such a pattern is less obvious in the plots of "weekday", "weathersit", "windspeed", and "holiday". This observation is consistent with the result that both the MBT and the Tree SHAP aggregation agree that the former 5 features are more important than the latter ones. Second, we observe that for the feature "hum", the points are scattered flatter than the diagonal line, while for the feature "workingday", the points are scattered steeper than the diagonal line. Such patterns are consistent with the results that "hum" is ranked higher in the MBT method, while "workingday" is ranked higher in Tree SHAP aggregation.

5. **Examine different subset selections in the MBT method**. We examine the feature importance results from the MBT method under different choices of feature subsets. We consider subset selection under $\gamma = 4$ (all subsets), $\gamma = 3$, $\gamma = 2$, $\gamma = 1$. The results are shown in Figure 16.

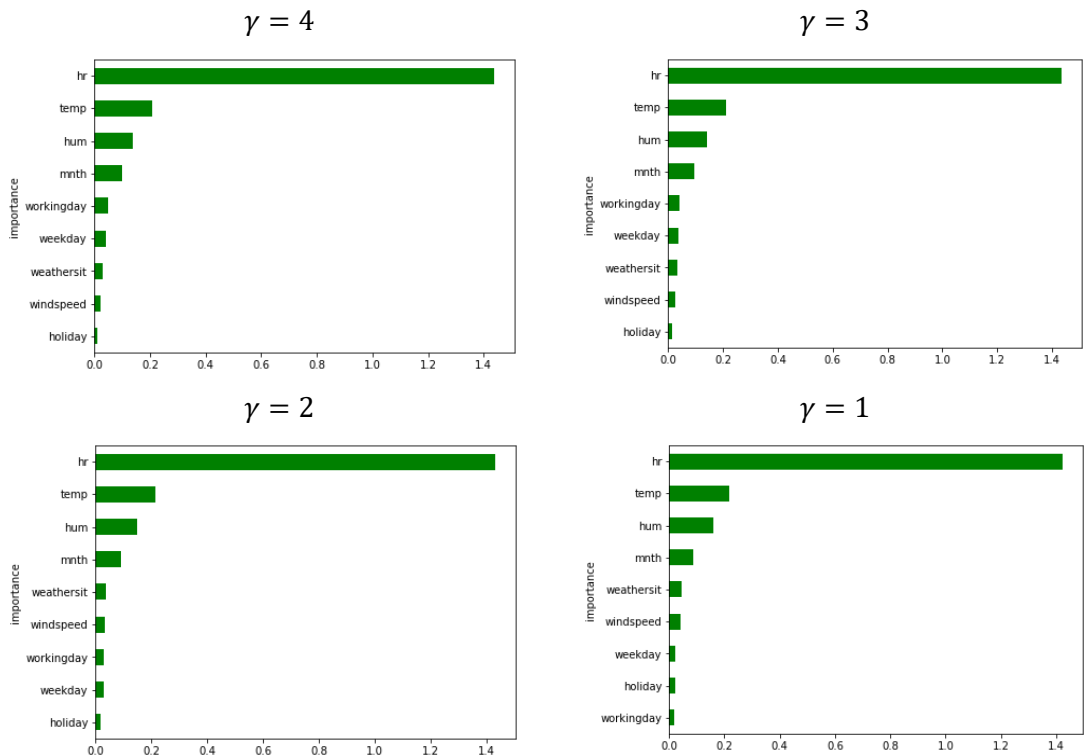

*Figure 16: Feature importance plot from the proposed algorithm under different choices of feature subsets.*

From Figure 16, we can see that the ranking results are relatively stable and do not change significantly, especially for the top 5 important features. Under different selections of subsets, the top 5 important features are consistently the same. The rankings of the last 4 features are the same when $\gamma$ decreases from 4 to 3. When $\gamma$ decreases further, there are some ranking swaps only for the features that are least important.

## 5 Summary and future work

In this paper, we proposed a method to compute the global Shapley values that use conditional distribution in the cost function. The proposed method relies on a model-based tree to compute the conditional expectation and offers a way to improve the computational efficiency. Various simulations studies have shown that, compared with major existing methods, the proposed method delivers remarkable improvement in computational accuracy, unifies both global and local interpretation under the same algorithm, and provides user the option to reduce running time at the expense of accuracy.

The proposed algorithm makes three major assumptions: 1) In each splitting node of the SLIM tree, only top important variables are used to compute the individual path probability; 2) In each terminal node, an additive model is assumed for local conditional expectation; 3) In pursuing computational efficiency, unselected subsets of features are ignored in the computation. The third assumption may lead to a major limitation. That is, the more computational efficiency is gained, the more accuracy we may lose. Therefore, while the proposed method is computational efficient for small and medium size of features, it is still limited for large size of features. For example, if the number of features is 100, then even one chooses $\gamma = 2$, there will still be about 10,000 subsets selected. In this case, one may be interested in feature elimination, namely eliminating strongly correlated features from the original dataset, then apply the thresholding method to the reduced dataset. Feature elimination may still be limited in many applications such as natural language processing, where the feature size is generally large. It is challenging to improve the efficiency for high-dimensional dataset and requires future work.

## 6 Acknowledgements

We are grateful to Wei Zhao who contributed to this research while she was working at Wells Fargo.

# Appendix 1

Empirical kernel estimation involves the following steps:

1. Given a feature subset $u$, compute the distance between an observation $x_u^*$ and all other observations $x_u^{(j)}$:

$$d\left(x_u^*, x_u^{(j)}\right) = \sqrt{\frac{\left(x_u^* - x_u^{(j)}\right)^T \Sigma_{u,u}^{-1} \left(x_u^* - x_u^{(j)}\right)}{|u|}}$$

2. Compute the Gaussian distribution kernel:

$$w_u\left(x_u^*, x_u^{(j)}\right) = exp\left(-\frac{d^2\left(x_u^*, x_u^{(j)}\right)}{2\varrho^2}\right)$$

with some value for the Gaussian kernel standard deviation $\varrho$.

3. Sort all kernels and select the first $k$ largest kernels and the corresponding observations $x^{(j)}$, and compute the conditional expectation as

$$\mathbb{E}(\hat{f}(X)|X_u = x_u) \approx \frac{\sum_{j=1}^{k} w_u\left(x_u^*, x_u^{(j)}\right) \hat{f}(x^{(j)})}{\sum_{j=1}^{k} w_u\left(x_u^*, x_u^{(j)}\right)}$$

4. Repeat step 1 – 3 for all observations to compute $Var(\mathbb{E}(\hat{f}(X)|X_u))$.
5. Repeat step 1 – 4 for all selected subsets $u$. Compute global Shapley.

This procedure is time-consuming and computationally inefficient. It requires to compute the Gaussian distribution kernel between each observation for all selected subsets, and the time complexity is $\mathcal{O}(|\mathcal{D}|n^3 \log(n))$, where $|\mathcal{D}|$ is the number of selected feature subsets and $n$ is the number of observations.

# Appendix 2

**Data preparation**

For the binary response case, the generated data are used to build an Xgboost model, which computes the classification probability $\hat{p}$ for each observation. Compute the log-odds ratios $\hat{\eta} = logit(\hat{p}) = ln\left(\frac{\hat{p}}{1-\hat{p}}\right)$. The input data $X$ and the corresponding log-odds $\hat{\eta}$ are used to compute the global Shapley, i.e., the global Shapley is computed in the log-odds space. The Xgboost classification tree model is well-tuned and the performance on the training and testing dataset, measured by AUC, is provided in Table 10.

| $\rho$ | 0.99 | 0.9 | 0.75 | 0.5 | 0.4 | 0.2 | 0 |
|---|---|---|---|---|---|---|---|
| AUC (train) | 0.988 | 0.989 | 0.987 | 0.984 | 0.983 | 0.98 | 0.973 |
| AUC (test) | 0.98 | 0.978 | 0.98 | 0.969 | 0.965 | 0.955 | 0.959 |

*Table 10: The performance of the Xgboost tree model using the training and test dataset.*

**Methods set-ups**

In the proposed algorithm, we select the SLIM tree depth is 1 for linear, nonlinear, and binary response cases, as these cases are GAM. For interaction case, SLIM tree depth is chosen as 5 after tuning. In each splitting node of a SLIM tree, we choose 3 top important variables for each splitting variable, i.e., $|v| = 3$. In each splitting node, we choose to build a random forest (RF) model to compute the individual path probability, each RF contains 10 estimators (trees) with maximum depth is 10, and the maximum feature size is chosen as $\sqrt{p}$. In each terminal node of a SLIM tree, a linear B-spline model with 10 knots is built to capture any nonlinear terms. The following table is the performance of the SLIM tree in different cases. The performance is measured by the Mean Squared Error (MSE) and the R-square (Rsq) using the training and test dataset, as shown in Table 11.

|  |  | MSE | | Rsq | |
| --- | --- | --- | --- | --- | --- |
|  |  | Train | Test | Train | Test |
| Linear Scenario | 0.99 | 0.308 | 0.286 | 0.994 | 0.995 |
|  | 0.9 | 0.237 | 0.222 | 0.995 | 0.996 |
|  | 0.75 | 0.178 | 0.165 | 0.996 | 0.996 |
|  | 0.5 | 0.154 | 0.148 | 0.996 | 0.996 |
|  | 0.4 | 0.152 | 0.15 | 0.995 | 0.996 |
|  | 0.2 | 0.15 | 0.141 | 0.994 | 0.995 |
|  | 0 | 0.148 | 0.151 | 0.993 | 0.993 |
| Non-linear Scenario | 0.99 | 0.168 | 0.204 | 0.997 | 0.996 |
|  | 0.9 | 0.146 | 0.200 | 0.997 | 0.996 |
|  | 0.75 | 0.116 | 0.208 | 0.997 | 0.995 |
|  | 0.5 | 0.120 | 0.146 | 0.997 | 0.996 |
|  | 0.4 | 0.115 | 0.233 | 0.996 | 0.993 |
|  | 0.2 | 0.120 | 0.177 | 0.996 | 0.994 |
|  | 0 | 0.116 | 0.187 | 0.995 | 0.991 |
| Interaction Scenario | 0.99 | 0.182 | 0.132 | 0.982 | 0.986 |
|  | 0.9 | 0.419 | 0.415 | 0.955 | 0.956 |
|  | 0.75 | 0.208 | 0.269 | 0.975 | 0.969 |
|  | 0.5 | 0.293 | 0.431 | 0.959 | 0.942 |
|  | 0.4 | 0.280 | 0.442 | 0.958 | 0.937 |
|  | 0.2 | 0.342 | 0.475 | 0.941 | 0.920 |
|  | 0 | 0.337 | 0.559 | 0.933 | 0.892 |
| Binary Response Scenario | 0.99 | 1.142 | 1.510 | 0.978 | 0.973 |
|  | 0.9 | 1.227 | 1.365 | 0.978 | 0.976 |
|  | 0.75 | 1.012 | 1.197 | 0.977 | 0.974 |
|  | 0.5 | 0.899 | 0.988 | 0.977 | 0.976 |
|  | 0.4 | 0.885 | 0.904 | 0.976 | 0.976 |
|  | 0.2 | 0.806 | 0.811 | 0.974 | 0.974 |
|  | 0 | 0.779 | 0.836 | 0.967 | 0.966 |

*Table 11: The performance of the SLIM tree in each simulation scenario and each correlation value.*

# Appendix 3

All the full results of the four simulation scenarios are presented in Table 12 - Table 15.

| ρ | | X1 | X2 | X3 | X4 | X5 | X6 | X7 | X8 | X9 | X10 | X11 | X12 | X13 | Error |
|---|---|---|---|---|---|---|---|---|---|---|---|---|---|---|---|
| | | | | | | | Linear Scenario | | | | | | | | |
| 0.99 | Oracle | 8.79 | 8.79 | 9.91 | 9.85 | 9.89 | 6.61 | 6.77 | 6.73 | 6.46 | 6.20 | 0.00 | 9.99 | 9.99 | - |
| | MBT | 8.75 | 8.76 | 9.72 | 9.67 | 9.71 | 6.61 | 6.77 | 6.76 | 6.51 | 6.30 | 0.01 | 10.23 | 10.21 | 0.00025 |
| | Marginal | 8.68 | 8.68 | 11.25 | 7.51 | 10.48 | 3.98 | 14.64 | 14.62 | 0.00 | 0.00 | 0.00 | 10.10 | 10.06 | 0.25374 |
| | Empirical | 8.68 | 10.25 | 8.92 | 7.93 | 7.13 | 7.09 | 6.89 | 5.94 | 5.85 | 7.72 | 8.04 | 7.72 | 7.83 | 0.10726 |
| | M.A.SHAP | 8.10 | 12.68 | 11.31 | 7.87 | 8.64 | 3.46 | 6.15 | 18.82 | 0.38 | 0.02 | 0.02 | 7.94 | 14.61 | 0.32458 |
| 0.9 | Oracle | 8.88 | 8.88 | 10.07 | 9.50 | 9.95 | 6.74 | 8.47 | 8.09 | 5.47 | 3.74 | 0.00 | 10.11 | 10.11 | - |
| | MBT | 9.14 | 9.03 | 9.92 | 9.36 | 9.82 | 6.67 | 8.51 | 8.04 | 5.49 | 3.87 | 0.01 | 10.06 | 10.08 | 0.00020 |
| | Marginal | 9.11 | 8.86 | 11.19 | 7.39 | 10.49 | 3.66 | 14.71 | 14.57 | 0.00 | 0.00 | 0.00 | 10.01 | 10.02 | 0.15966 |
| | Empirical | 2.94 | 10.81 | 7.75 | 11.58 | 5.75 | 10.50 | 7.08 | 8.55 | 14.95 | 4.13 | 7.14 | 4.05 | 4.78 | 0.32842 |
| | M.A.SHAP | 11.68 | 9.57 | 11.84 | 6.21 | 9.30 | 2.57 | 12.85 | 13.26 | 0.09 | 0.04 | 0.02 | 12.06 | 10.51 | 0.15080 |
| 0.75 | Oracle | 9.05 | 9.05 | 10.32 | 8.88 | 10.02 | 6.07 | 10.67 | 9.86 | 3.92 | 1.56 | 0.00 | 10.30 | 10.30 | - |
| | MBT | 9.31 | 9.17 | 10.15 | 8.69 | 9.86 | 6.18 | 10.64 | 10.05 | 3.91 | 1.54 | 0.00 | 10.21 | 10.29 | 0.00024 |
| | Marginal | 9.36 | 8.97 | 11.11 | 7.22 | 10.40 | 3.22 | 14.69 | 14.56 | 0.00 | 0.00 | 0.00 | 10.24 | 10.22 | 0.07314 |
| | Empirical | 1.81 | 13.24 | 6.51 | 14.02 | 5.49 | 4.52 | 16.11 | 6.33 | 9.97 | 7.91 | 6.33 | 3.82 | 3.94 | 0.40470 |
| | M.A.SHAP | 10.49 | 10.81 | 10.31 | 7.07 | 10.13 | 3.18 | 13.56 | 11.96 | 0.12 | 0.04 | 0.10 | 11.16 | 11.07 | 0.05137 |
| 0.5 | Oracle | 9.39 | 9.39 | 10.66 | 7.71 | 10.02 | 4.01 | 13.21 | 12.17 | 1.81 | 0.27 | 0.00 | 10.68 | 10.68 | - |
| | MBT | 9.47 | 9.23 | 10.41 | 7.63 | 9.94 | 3.97 | 13.01 | 12.38 | 1.85 | 0.26 | 0.00 | 10.85 | 11.01 | 0.00033 |
| | Marginal | 9.62 | 9.11 | 11.07 | 6.86 | 10.17 | 2.56 | 14.50 | 14.36 | 0.00 | 0.00 | 0.00 | 10.90 | 10.86 | 0.01280 |
| | Empirical | 4.29 | 6.75 | 6.60 | 9.12 | 6.84 | 8.01 | 15.41 | 6.38 | 3.38 | 9.52 | 7.66 | 9.53 | 6.50 | 0.27585 |
| | M.A.SHAP | 10.41 | 10.47 | 10.68 | 6.99 | 10.07 | 3.55 | 13.17 | 12.28 | 0.02 | 0.12 | 0.02 | 11.06 | 11.16 | 0.00645 |
| 0.4 | Oracle | 9.54 | 9.54 | 10.75 | 7.19 | 9.97 | 3.22 | 13.89 | 12.91 | 1.17 | 0.11 | 0.00 | 10.86 | 10.86 | - |
| | MBT | 9.68 | 9.39 | 10.48 | 7.02 | 9.87 | 3.17 | 14.18 | 12.86 | 1.14 | 0.07 | 0.00 | 11.04 | 11.10 | 0.00032 |
| | Marginal | 9.94 | 9.62 | 10.84 | 6.54 | 10.21 | 2.26 | 14.72 | 13.94 | 0.00 | 0.00 | 0.00 | 11.08 | 10.84 | 0.00486 |
| | Empirical | 4.31 | 7.24 | 7.67 | 12.08 | 6.76 | 13.23 | 5.70 | 6.50 | 7.38 | 3.82 | 8.61 | 10.19 | 6.52 | 0.40764 |
| | M.A.SHAP | 10.49 | 10.34 | 10.73 | 7.11 | 9.79 | 3.25 | 12.93 | 12.59 | 0.06 | 0.07 | 0.03 | 11.20 | 11.42 | 0.00404 |
| 0.2 | Oracle | 9.92 | 9.92 | 10.77 | 6.01 | 9.71 | 1.94 | 14.74 | 14.09 | 0.30 | 0.01 | 0.00 | 11.29 | 11.29 | - |
| | MBT | 10.21 | 10.03 | 10.92 | 5.91 | 9.82 | 1.81 | 14.80 | 13.83 | 0.23 | 0.00 | 0.00 | 11.46 | 11.00 | 0.00031 |
| | Marginal | 10.29 | 9.77 | 10.69 | 5.83 | 9.88 | 1.66 | 14.70 | 13.99 | 0.00 | 0.00 | 0.00 | 11.79 | 11.38 | 0.00486 |
| | Empirical | 10.76 | 8.72 | 5.90 | 8.08 | 5.62 | 11.71 | 7.84 | 7.51 | 6.46 | 6.20 | 8.53 | 5.09 | 7.57 | 0.38878 |
| | M.A.SHAP | 10.45 | 10.48 | 10.53 | 7.09 | 9.94 | 3.37 | 13.08 | 12.57 | 0.01 | 0.06 | 0.01 | 11.09 | 11.31 | 0.00818 |
| 0 | Oracle | 10.44 | 10.44 | 10.44 | 4.64 | 9.09 | 1.16 | 15.03 | 15.03 | 0.00 | 0.00 | 0.00 | 11.87 | 11.87 | - |
| | MBT | 10.38 | 10.08 | 10.34 | 4.90 | 8.86 | 1.12 | 14.73 | 15.26 | 0.00 | 0.00 | 0.00 | 11.84 | 12.50 | 0.00068 |
| | Marginal | 11.18 | 10.05 | 10.46 | 4.73 | 8.95 | 1.09 | 14.51 | 15.13 | 0.00 | 0.00 | 0.00 | 11.70 | 12.20 | 0.00098 |
| | Empirical | 11.15 | 6.90 | 5.26 | 4.57 | 17.80 | 2.42 | 6.36 | 17.77 | 4.98 | 5.86 | 8.33 | 4.59 | 4.02 | 0.38012 |
| | M.A.SHAP | 10.52 | 10.77 | 10.68 | 5.52 | 9.64 | 0.01 | 13.96 | 14.18 | 0.00 | 0.00 | 0.00 | 12.28 | 12.44 | 0.00422 |

*Table 12: Full results for linear scenario.*

| $\rho$ | | X1 | X2 | X3 | X4 | X5 | X6 | X7 | X8 | X9 | X10 | X11 | X12 | X13 | Error |
|---|---|---|---|---|---|---|---|---|---|---|---|---|---|---|---|
| | | | | | | **Non-linear Scenario** | | | | | | | | | |
| 0.99 | Oracle | 8.79 | 8.79 | 9.91 | 9.85 | 9.90 | 6.61 | 6.77 | 6.73 | 6.45 | 6.20 | 0.00 | 10.00 | 10.00 | - |
| | MBT | 8.76 | 8.75 | 9.74 | 9.67 | 9.70 | 6.61 | 6.77 | 6.75 | 6.52 | 6.30 | 0.01 | 10.23 | 10.19 | 0.00024 |
| | Marginal | 8.69 | 8.69 | 11.25 | 7.50 | 10.48 | 3.98 | 14.64 | 14.62 | 0.00 | 0.00 | 0.00 | 10.10 | 10.06 | 0.25367 |
| | Empirical | 7.72 | 8.58 | 4.13 | 6.77 | 9.86 | 7.88 | 8.73 | 8.00 | 8.12 | 10.74 | 6.81 | 6.59 | 6.06 | 0.17153 |
| | M.A.SHAP | 8.14 | 12.58 | 11.29 | 8.18 | 8.50 | 3.82 | 6.11 | 18.51 | 0.34 | 0.02 | 0.02 | 8.07 | 14.43 | 0.30998 |
| 0.9 | Oracle | 8.88 | 8.88 | 10.07 | 9.50 | 9.95 | 6.74 | 8.47 | 8.09 | 5.47 | 3.74 | 0.00 | 10.11 | 10.11 | - |
| | MBT | 9.14 | 9.02 | 9.93 | 9.37 | 9.82 | 6.68 | 8.51 | 8.04 | 5.49 | 3.87 | 0.01 | 10.06 | 10.07 | 0.00019 |
| | Marginal | 9.11 | 8.86 | 11.19 | 7.39 | 10.49 | 3.66 | 14.71 | 14.57 | 0.00 | 0.00 | 0.00 | 10.01 | 10.02 | 0.15966 |
| | Empirical | 3.10 | 8.86 | 4.13 | 10.78 | 7.29 | 11.07 | 11.47 | 6.56 | 15.57 | 3.59 | 5.88 | 4.02 | 7.66 | 0.32651 |
| | M.A.SHAP | 11.86 | 9.31 | 11.91 | 6.14 | 9.22 | 2.54 | 12.81 | 13.42 | 0.11 | 0.07 | 0.08 | 12.13 | 10.41 | 0.15407 |
| 0.75 | Oracle | 9.05 | 9.05 | 10.32 | 8.88 | 10.02 | 6.07 | 10.67 | 9.86 | 3.92 | 1.56 | 0.00 | 10.30 | 10.30 | - |
| | MBT | 9.32 | 9.16 | 10.17 | 8.69 | 9.85 | 6.19 | 10.63 | 10.04 | 3.91 | 1.54 | 0.00 | 10.21 | 10.29 | 0.00024 |
| | Marginal | 9.36 | 8.97 | 11.11 | 7.22 | 10.40 | 3.22 | 14.69 | 14.56 | 0.00 | 0.00 | 0.00 | 10.24 | 10.22 | 0.07314 |
| | Empirical | 2.45 | 12.78 | 4.50 | 12.59 | 4.73 | 3.79 | 20.82 | 5.37 | 13.07 | 4.29 | 5.65 | 3.31 | 6.65 | 0.48188 |
| | M.A.SHAP | 10.50 | 10.75 | 10.11 | 7.06 | 10.46 | 3.05 | 13.55 | 12.13 | 0.00 | 0.06 | 0.07 | 10.98 | 11.26 | 0.05400 |
| 0.5 | Oracle | 9.39 | 9.39 | 10.66 | 7.71 | 10.02 | 4.01 | 13.21 | 12.17 | 1.81 | 0.27 | 0.00 | 10.68 | 10.68 | - |
| | MBT | 9.47 | 9.22 | 10.42 | 7.62 | 9.94 | 3.97 | 13.00 | 12.38 | 1.85 | 0.26 | 0.00 | 10.85 | 11.01 | 0.00033 |
| | Marginal | 9.62 | 9.11 | 11.07 | 6.86 | 10.17 | 2.56 | 14.50 | 14.36 | 0.00 | 0.00 | 0.00 | 10.90 | 10.86 | 0.01280 |
| | Empirical | 4.71 | 7.26 | 5.81 | 8.56 | 7.35 | 5.87 | 18.21 | 6.62 | 1.76 | 9.19 | 7.21 | 7.04 | 10.42 | 0.25687 |
| | M.A.SHAP | 10.43 | 10.45 | 10.71 | 7.01 | 10.06 | 3.59 | 13.12 | 12.33 | 0.02 | 0.12 | 0.02 | 11.06 | 11.08 | 0.00633 |
| 0.4 | Oracle | 9.54 | 9.54 | 10.75 | 7.19 | 9.97 | 3.22 | 13.89 | 12.91 | 1.17 | 0.11 | 0.00 | 10.86 | 10.86 | - |
| | MBT | 9.69 | 9.29 | 10.49 | 7.01 | 9.87 | 3.18 | 14.17 | 12.92 | 1.13 | 0.07 | 0.00 | 11.06 | 11.10 | 0.00036 |
| | Marginal | 9.94 | 9.62 | 10.84 | 6.54 | 10.21 | 2.26 | 14.72 | 13.94 | 0.00 | 0.00 | 0.00 | 11.08 | 10.84 | 0.00097 |
| | Empirical | 3.97 | 6.79 | 5.72 | 15.69 | 6.04 | 14.58 | 7.28 | 4.86 | 5.47 | 3.08 | 6.92 | 8.83 | 10.75 | 0.44377 |
| | M.A.SHAP | 10.56 | 10.33 | 10.74 | 7.09 | 9.76 | 3.24 | 12.89 | 12.65 | 0.05 | 0.03 | 0.03 | 11.16 | 11.47 | 0.00428 |
| 0.2 | Oracle | 9.92 | 9.92 | 10.77 | 6.01 | 9.71 | 1.94 | 14.74 | 14.09 | 0.30 | 0.01 | 0.00 | 11.29 | 11.29 | - |
| | MBT | 10.21 | 10.02 | 10.93 | 5.92 | 9.81 | 1.81 | 14.80 | 13.83 | 0.23 | 0.00 | 0.00 | 11.45 | 11.00 | 0.00030 |
| | Marginal | 10.29 | 9.77 | 10.69 | 5.83 | 9.88 | 1.66 | 14.72 | 13.99 | 0.00 | 0.00 | 0.00 | 11.79 | 11.38 | 0.00059 |
| | Empirical | 10.58 | 9.30 | 6.12 | 7.32 | 5.35 | 11.77 | 7.92 | 7.75 | 5.68 | 6.18 | 9.00 | 5.03 | 7.99 | 0.37985 |
| | M.A.SHAP | 10.57 | 10.43 | 10.79 | 6.91 | 10.03 | 2.51 | 13.35 | 12.92 | 0.00 | 0.00 | 0.00 | 11.29 | 11.19 | 0.00475 |
| 0 | Oracle | 10.44 | 10.44 | 10.44 | 4.64 | 9.09 | 1.16 | 15.03 | 15.03 | 0.00 | 0.00 | 0.00 | 11.87 | 11.87 | - |
| | MBT | 10.37 | 10.07 | 10.35 | 4.89 | 8.86 | 1.13 | 14.67 | 15.33 | 0.00 | 0.00 | 0.00 | 11.84 | 12.49 | 0.00073 |
| | Marginal | 11.18 | 10.05 | 10.46 | 4.73 | 8.95 | 1.09 | 14.51 | 15.13 | 0.00 | 0.00 | 0.00 | 11.70 | 12.20 | 0.00098 |
| | Empirical | 9.63 | 7.67 | 4.54 | 4.51 | 14.62 | 6.99 | 8.93 | 17.00 | 5.45 | 4.20 | 8.20 | 4.51 | 3.75 | 0.32890 |
| | M.A.SHAP | 10.52 | 10.62 | 10.61 | 6.34 | 9.84 | 2.33 | 13.18 | 13.22 | 0.00 | 0.00 | 0.00 | 11.60 | 11.74 | 0.00999 |

*Table 13: Full results for non-linear scenario.*

| ρ | | X1 | X2 | X3 | X4 | X5 | X6 | X7 | X8 | X9 | X10 | X11 | X12 | X13 | Error |
|---|---|---|---|---|---|---|---|---|---|---|---|---|---|---|---|
| | | | | | | | **Binary Response Scenario** | | | | | | | | |
| 0.99 | Oracle | 8.79 | 8.79 | 9.91 | 9.85 | 9.90 | 6.61 | 6.77 | 6.73 | 6.45 | 6.20 | 0.00 | 10.00 | 10.00 | - |
| | MBT | 9.16 | 9.20 | 9.81 | 9.65 | 9.65 | 6.49 | 6.56 | 6.54 | 6.30 | 6.04 | 0.10 | 10.24 | 10.26 | 0.00081 |
| | Marginal | 8.69 | 8.68 | 11.25 | 7.50 | 10.48 | 3.98 | 14.64 | 14.62 | 0.00 | 0.00 | 0.00 | 10.10 | 10.06 | 0.25367 |
| | Empirical | 8.96 | 9.35 | 4.13 | 5.62 | 7.26 | 8.07 | 8.06 | 12.20 | 9.37 | 9.24 | 7.54 | 5.35 | 4.85 | 0.24926 |
| | M.A.SHAP | 11.15 | 8.09 | 11.64 | 5.82 | 10.86 | 2.90 | 17.08 | 5.77 | 4.05 | 0.99 | 0.44 | 15.04 | 6.17 | 0.25550 |
| 0.9 | Oracle | 8.88 | 8.88 | 10.07 | 9.50 | 9.95 | 6.74 | 8.47 | 8.09 | 5.47 | 3.74 | 0.00 | 10.11 | 10.11 | - |
| | MBT | 9.33 | 9.39 | 9.81 | 9.06 | 9.63 | 6.49 | 8.18 | 8.04 | 5.38 | 3.86 | 0.12 | 10.27 | 10.44 | 0.00131 |
| | Marginal | 9.11 | 8.86 | 11.19 | 7.39 | 10.49 | 3.66 | 14.71 | 14.57 | 0.00 | 0.00 | 0.00 | 10.01 | 10.02 | 0.15966 |
| | Empirical | 3.30 | 8.88 | 4.23 | 9.20 | 6.98 | 9.93 | 12.58 | 7.35 | 17.24 | 3.35 | 5.78 | 3.78 | 7.40 | 0.36565 |
| | M.A.SHAP | 10.52 | 9.14 | 9.87 | 6.45 | 11.07 | 1.97 | 14.34 | 13.08 | 0.99 | 0.40 | 0.51 | 10.60 | 11.07 | 0.14572 |
| 0.75 | Oracle | 9.05 | 9.05 | 10.32 | 8.88 | 10.02 | 6.07 | 10.67 | 9.86 | 3.82 | 1.56 | 0.00 | 10.30 | 10.30 | - |
| | MBT | 9.02 | 9.33 | 9.85 | 8.96 | 9.66 | 6.41 | 10.83 | 9.50 | 4.06 | 1.76 | 0.10 | 10.05 | 10.48 | 0.00095 |
| | Marginal | 9.36 | 8.97 | 11.11 | 7.22 | 10.40 | 3.20 | 14.69 | 14.56 | 0.00 | 0.00 | 0.00 | 10.24 | 10.22 | 0.07314 |
| | Empirical | 2.11 | 12.25 | 4.56 | 12.15 | 4.75 | 3.67 | 20.66 | 5.42 | 13.77 | 4.29 | 5.39 | 3.44 | 7.54 | 0.47836 |
| | M.A.SHAP | 9.72 | 10.31 | 11.28 | 6.24 | 9.87 | 2.97 | 12.78 | 13.46 | 0.48 | 0.85 | 0.56 | 10.79 | 10.68 | 0.05391 |
| 0.5 | Oracle | 9.39 | 9.39 | 10.66 | 7.71 | 10.02 | 4.01 | 13.21 | 12.17 | 1.81 | 0.27 | 0.00 | 10.68 | 10.68 | - |
| | MBT | 10.06 | 9.55 | 10.00 | 7.21 | 9.86 | 3.90 | 12.39 | 12.68 | 1.86 | 0.37 | 0.07 | 11.00 | 11.05 | 0.00233 |
| | Marginal | 9.62 | 9.11 | 11.07 | 6.86 | 10.17 | 2.56 | 14.50 | 14.36 | 0.00 | 0.00 | 0.00 | 10.90 | 10.86 | 0.01280 |
| | Empirical | 4.84 | 7.76 | 6.05 | 7.19 | 6.88 | 6.01 | 18.05 | 6.79 | 1.69 | 9.26 | 7.65 | 7.39 | 10.44 | 0.25617 |
| | M.A.SHAP | 9.85 | 9.91 | 11.10 | 6.89 | 9.84 | 3.88 | 12.95 | 12.49 | 0.25 | 0.47 | 0.88 | 10.70 | 10.79 | 0.00476 |
| 0.4 | Oracle | 9.54 | 9.54 | 10.75 | 7.19 | 9.97 | 3.22 | 13.89 | 12.91 | 1.17 | 0.11 | 0.00 | 10.86 | 10.86 | - |
| | MBT | 9.14 | 9.68 | 9.96 | 7.20 | 9.56 | 3.37 | 14.07 | 12.59 | 1.48 | 0.19 | 0.14 | 11.05 | 11.57 | 0.00170 |
| | Marginal | 9.86 | 9.28 | 10.92 | 6.58 | 10.03 | 2.28 | 14.85 | 14.06 | 0.01 | 0.01 | 0.01 | 11.13 | 10.96 | 0.00486 |
| | Empirical | 4.04 | 6.60 | 5.39 | 17.17 | 5.91 | 14.74 | 7.05 | 5.24 | 5.18 | 3.37 | 5.75 | 9.24 | 10.32 | 0.45909 |
| | M.A.SHAP | 10.09 | 10.16 | 10.94 | 6.46 | 9.95 | 3.80 | 13.22 | 12.06 | 0.75 | 0.54 | 0.50 | 10.91 | 10.62 | 0.00323 |
| 0.2 | Oracle | 9.92 | 9.92 | 10.77 | 6.01 | 9.71 | 1.94 | 14.74 | 14.09 | 0.30 | 0.01 | 0.00 | 11.29 | 11.29 | - |
| | MBT | 9.66 | 10.10 | 11.30 | 6.17 | 9.82 | 1.86 | 14.30 | 13.94 | 0.39 | 0.11 | 0.09 | 10.96 | 11.31 | 0.00070 |
| | Marginal | 10.29 | 9.95 | 10.89 | 6.02 | 9.84 | 1.67 | 14.68 | 13.89 | 0.02 | 0.02 | 0.02 | 11.72 | 10.97 | 0.00058 |
| | Empirical | 9.48 | 9.63 | 6.23 | 7.32 | 5.10 | 10.63 | 9.35 | 7.45 | 5.84 | 5.60 | 10.85 | 4.81 | 7.71 | 0.38218 |
| | M.A.SHAP | 10.16 | 10.39 | 10.31 | 6.76 | 9.89 | 3.66 | 12.80 | 12.89 | 0.50 | 0.55 | 0.59 | 10.83 | 10.68 | 0.00943 |
| 0 | Oracle | 10.44 | 10.44 | 10.44 | 4.64 | 9.09 | 1.16 | 15.03 | 15.03 | 0.00 | 0.00 | 0.00 | 11.87 | 11.87 | - |
| | MBT | 10.36 | 10.75 | 9.59 | 4.42 | 8.83 | 1.22 | 14.68 | 15.01 | 0.13 | 0.14 | 0.14 | 12.35 | 12.37 | 0.00136 |
| | Marginal | 11.18 | 10.05 | 10.46 | 4.73 | 8.95 | 1.09 | 14.51 | 15.13 | 0.00 | 0.00 | 0.00 | 11.70 | 12.20 | 0.00098 |
| | Empirical | 9.83 | 8.03 | 4.17 | 4.27 | 16.00 | 7.74 | 7.85 | 16.35 | 5.41 | 4.34 | 9.02 | 4.31 | 2.67 | 0.39564 |
| | M.A.SHAP | 10.20 | 10.54 | 9.83 | 7.35 | 9.68 | 3.76 | 12.89 | 12.36 | 0.32 | 0.43 | 0.37 | 11.05 | 11.21 | 0.02413 |

*Table 14: Full results for binary response scenario.*

|  |  | Interaction Scenario | | | | | | |
| --- | --- | --- | --- | --- | --- | --- | --- | --- |
|  |  | X1 | X2 | X3 | X4 | X5 | X6 | Error |
| 0.99 | Oracle | 31.10 | 32.31 | 30.57 | 2.93 | 3.08 | 0.00 | - |
|  | MBT | 30.69 | 32.14 | 30.28 | 3.19 | 3.37 | 0.32 | 0.00018 |
|  | Marginal | 29.57 | 60.98 | 4.95 | 0.00 | 4.50 | 0.00 | 0.50298 |
|  | Empirical | 17.80 | 20.44 | 17.48 | 13.57 | 15.62 | 15.08 | 0.33293 |
|  | M.A.SHAP | 26.30 | 59.36 | 1.27 | 0.90 | 12.13 | 0.05 | 0.57317 |
| 0.9 | Oracle | 29.09 | 40.30 | 24.43 | 2.38 | 3.80 | 0.00 | - |
|  | MBT | 28.07 | 38.88 | 23.74 | 2.82 | 5.64 | 0.84 | 0.00253 |
|  | Marginal | 29.69 | 60.72 | 4.96 | 0.00 | 4.64 | 0.00 | 0.25986 |
|  | Empirical | 20.71 | 37.40 | 9.63 | 8.44 | 12.30 | 11.52 | 0.17470 |
|  | M.A.SHAP | 29.97 | 52.30 | 5.87 | 0.32 | 11.48 | 0.06 | 0.17895 |
| 0.75 | Oracle | 26.25 | 50.37 | 16.86 | 1.61 | 4.91 | 0.00 | - |
|  | MBT | 25.42 | 50.04 | 17.32 | 1.90 | 4.92 | 0.39 | 0.00035 |
|  | Marginal | 29.48 | 59.99 | 5.36 | 0.00 | 5.18 | 0.00 | 0.06726 |
|  | Empirical | 11.39 | 42.87 | 18.43 | 4.67 | 10.57 | 12.07 | 0.13193 |
|  | M.A.SHAP | 29.08 | 51.21 | 7.80 | 0.85 | 11.00 | 0.06 | 0.03635 |
| 0.5 | Oracle | 23.74 | 58.71 | 10.26 | 0.70 | 6.59 | 0.00 | - |
|  | MBT | 23.80 | 57.22 | 10.53 | 1.32 | 6.45 | 0.68 | 0.00075 |
|  | Marginal | 29.18 | 58.09 | 6.42 | 0.00 | 6.31 | 0.00 | 0.01087 |
|  | Empirical | 7.00 | 50.89 | 12.22 | 7.79 | 10.09 | 12.01 | 0.13271 |
|  | M.A.SHAP | 30.46 | 47.12 | 10.57 | 0.21 | 11.57 | 0.07 | 0.04914 |
| 0.4 | Oracle | 23.76 | 59.35 | 9.21 | 0.45 | 7.23 | 0.00 | - |
|  | MBT | 24.74 | 57.09 | 9.44 | 0.98 | 7.05 | 0.71 | 0.00164 |
|  | Marginal | 29.12 | 57.07 | 6.97 | 0.00 | 6.84 | 0.00 | 0.00929 |
|  | Empirical | 6.66 | 49.58 | 11.06 | 9.36 | 10.34 | 12.99 | 0.15365 |
|  | M.A.SHAP | 30.18 | 47.00 | 10.98 | 0.33 | 11.43 | 0.08 | 0.05075 |
| 0.2 | Oracle | 25.78 | 56.68 | 8.88 | 0.12 | 8.54 | 0.00 | - |
|  | MBT | 26.21 | 53.72 | 9.13 | 1.09 | 8.90 | 0.94 | 0.00272 |
|  | Marginal | 28.90 | 54.41 | 8.55 | 0.00 | 8.14 | 0.00 | 0.00375 |
|  | Empirical | 22.69 | 21.71 | 16.55 | 13.49 | 6.86 | 18.70 | 0.45232 |
|  | M.A.SHAP | 31.20 | 44.41 | 11.97 | 0.07 | 12.12 | 0.22 | 0.05021 |
| 0 | Oracle | 30.00 | 50.00 | 10.00 | 0.00 | 10.00 | 0.00 | - |
|  | MBT | 30.43 | 47.29 | 10.12 | 1.07 | 10.06 | 1.03 | 0.00270 |
|  | Marginal | 31.01 | 48.65 | 10.74 | 0.00 | 9.60 | 0.00 | 0.00098 |
|  | Empirical | 19.13 | 20.82 | 17.05 | 16.50 | 13.19 | 13.33 | 0.41099 |
|  | M.A.SHAP | 32.89 | 41.21 | 12.86 | 0.07 | 12.88 | 0.09 | 0.02838 |

*Table 15: Full results for interaction scenario.*